\documentclass[11pt,a4paper]{article}
\usepackage[hyperref]{emnlp2020}
\usepackage{times}
\usepackage{latexsym}

\usepackage{microtype}

\usepackage{amsmath,amsfonts,bm}

\def\eqref#1{equation~\ref{#1}}

\def\1{\bm{1}}

\DeclareMathAlphabet{\mathsfit}{\encodingdefault}{\sfdefault}{m}{sl}
\SetMathAlphabet{\mathsfit}{bold}{\encodingdefault}{\sfdefault}{bx}{n}

\def\gD{{\mathcal{D}}}

\def\gL{{\mathcal{L}}}

\def\gU{{\mathcal{U}}}

\newcommand{\E}{\mathbb{E}}

\newcommand{\R}{\mathbb{R}}

\usepackage{hyperref}
\usepackage{graphicx}
\usepackage{caption}
\usepackage{subcaption}
\usepackage{physics}
\usepackage{enumitem}
\usepackage{booktabs}
\usepackage{mathtools}
\usepackage{footmisc}

\makeatletter
\def\blfootnote{\xdef\@thefnmark{}\@footnotetext}
\makeatother

\aclfinalcopy
\title{How do Decisions Emerge across Layers in Neural Models? \\  Interpretation with Differentiable Masking}

\author{Nicola De Cao~\textsuperscript{1,2}, Michael Schlichtkrull~\textsuperscript{1,2}, Wilker Aziz~\textsuperscript{1}, Ivan Titov~\textsuperscript{1,2} \\
\textsuperscript{1}University of Amsterdam,
\textsuperscript{2}University of Edinburgh \\
{\tt \{ nicola.decao, m.s.schlichtkrull, w.aziz \} @uva.nl }\\
{\tt ititov@inf.ed.ac.uk}}

\date{}

\begin{document}
\maketitle

\begin{abstract}
Attribution methods assess the contribution of inputs to the model prediction. One way to do so is {\it erasure}: a subset of inputs is considered irrelevant if it can be removed without affecting the prediction. Though conceptually simple, erasure's objective is intractable and approximate search remains expensive with modern deep NLP models. Erasure is also susceptible to the {\it hindsight bias}: the fact that an input can be dropped does not mean that the model `knows' it can be dropped. The resulting pruning is over-aggressive and does not reflect how the model arrives at the prediction. To deal with these challenges, we introduce Differentiable Masking. \textsc{DiffMask} learns to mask-out subsets of the input while maintaining differentiability. The decision to include or disregard an input token is made with a simple model based on intermediate hidden layers of the analyzed model. First, this makes the approach efficient because we predict rather than search. Second, as with probing classifiers, this reveals what the network `knows' at the corresponding layers. This lets us not only plot attribution heatmaps but also analyze how decisions are formed across network layers. We use \textsc{DiffMask} to study BERT models on sentiment classification and question answering.\footnote{Source code available at \url{https://github.com/nicola-decao/diffmask}}
\end{abstract}

\begin{figure}[t]
	\centering
	\includegraphics[width=0.45\textwidth]{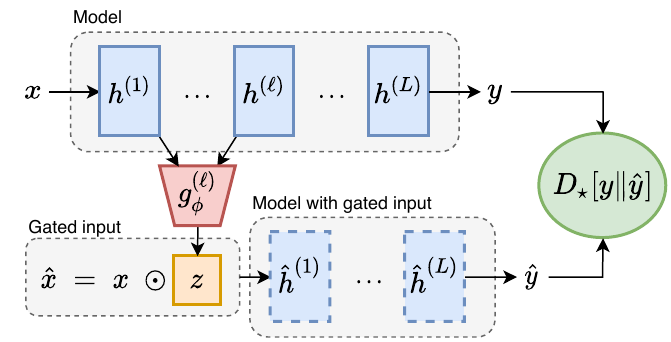}
	\caption{\textsc{DiffMask}: hidden states up to layer $\ell$ from a model (top) are fed to a classifier $g$ that predicts a mask $z$. We use this to mask the input and re-compute the forward pass (bottom). The classifier $g$ is trained to mask the input as much as possible without changing the output (minimizing a divergence $\mathrm{D_\star}$).}
	\label{fig:model-input}
\end{figure}

\begin{figure*}[t]
	\centering
	\begin{subfigure}[b]{0.49\textwidth}
		\centering
		\includegraphics[scale=0.9]{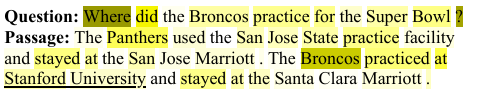}
		\caption{Integrated Gradient~\citep{sundararajan2017axiomatic}.}
		\label{fig:squad-comparison-ig}
	\end{subfigure}
	~
	\begin{subfigure}[b]{0.49\textwidth}
		\centering
		\includegraphics[scale=0.9]{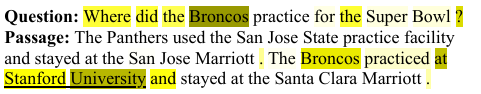}
		\caption{Restricting the Flow~\citep{schulz2020restricting}}
		\label{fig:squad-comparison-schulz}
	\end{subfigure}
    \par\vspace{6pt}
	\begin{subfigure}[b]{0.49\textwidth}
		\centering
		\includegraphics[scale=0.9]{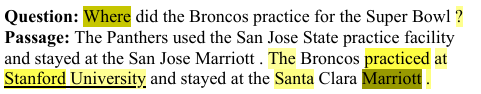}
		\caption{NLP explainer~\citep{guan2019towards}.}
		\label{fig:squad-comparison-guan}
	\end{subfigure}
	~
	\begin{subfigure}[b]{0.49\textwidth}
		\centering
		\includegraphics[scale=0.9]{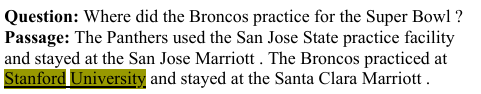}
		\caption{Erasure exact search optima.}
		\label{fig:squad-comparison-exact}
	\end{subfigure}
    \par\vspace{6pt}
	\begin{subfigure}[b]{0.49\textwidth}
		\centering
		\includegraphics[scale=0.9]{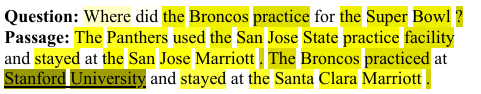}
		\caption{Our \textsc{DiffMask}.}
		\label{fig:squad-comparison-ours}
	\end{subfigure}
	~
	\begin{subfigure}[b]{0.49\textwidth}
		\centering
		\includegraphics[scale=0.9]{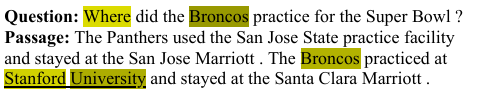}
		\caption{Our \textsc{DiffMask} non-amortized.}
		\label{fig:squad-comparison-ours-nonamo}
	\end{subfigure}
	\caption{Question Answering token attribution: (b) and (c), are misleading (i.e., not faithful) as they attribute the prediction mostly to the answer span itself (underlined). Our method (d) reveals that the model pays attention to other named entities and the predicate `practice' in both sentences. Predictions of the path-based methods (a) are more spread-out. Exact search (e) as well as approximate search (f) leads to pathological attributions.}
	\label{fig:squad-comparison}
\end{figure*}

\section{Introduction}

Deep neural networks have become standard tools in NLP demonstrating impressive improvements over traditional approaches on many tasks~\citep{goldberg2017neural}. Their power typically comes at the expense of interpretability, which may prevent users from trusting predictions~\cite{kim2015interactive,ribeiro2016should},  makes it hard to detect model or data deficiencies~\cite{gururangan2018annotation,kaushik2018much} or verify that a model is fair and does not exhibit harmful biases~\cite{sun2019mitigating,holstein2019improving}.  

These challenges have motivated work on interpretability, both in NLP and generally in machine learning; see~\citet{belinkov2019analysis} and~\citet{jacovi2020towards} for reviews. In this work, we study {\it post hoc interpretability} where the goal is to explain the prediction of a trained model and to reveal how the model arrives at the decision. This goal is usually approached with attribution methods~\citep{bach2015pixel, shrikumar2017learning, sundararajan2017axiomatic}, which explain the behavior of a model by assigning relevance to inputs.

One way to perform attribution is to use {\it erasure} where a subset of features (e.g., input tokens) is considered irrelevant if it can be removed without affecting the model prediction~\citep{li2016understanding,feng2018pathologies}. The advantage of erasure is that it is conceptually simple and optimizes a well-defined objective. This contrasts with most other attribution methods which rely on heuristic rules to define feature salience; for example, attention-based attribution~\citep{rocktaschel2015reasoning,serrano2019attention,vashishth2019attention} or back-propagation methods~\citep{bach2015pixel,shrikumar2017learning,sundararajan2017axiomatic}. These approaches received much scrutiny in recent years~\citep{nie2018theoretical,sixt2019explanations,jain2019attention}, as they cannot guarantee that the network is ignoring low-scored features. They are often motivated as approximations of erasure~\citep{baehrens2010explain,simonyan2013deep,feng2018pathologies} and sometimes  evaluated using erasure as ground-truth~\citep{serrano2019attention,jain2019attention}.

\begin{figure}[t]
	\centering
	\begin{subfigure}[b]{0.23\textwidth}
		\centering
		\includegraphics[scale=.2]{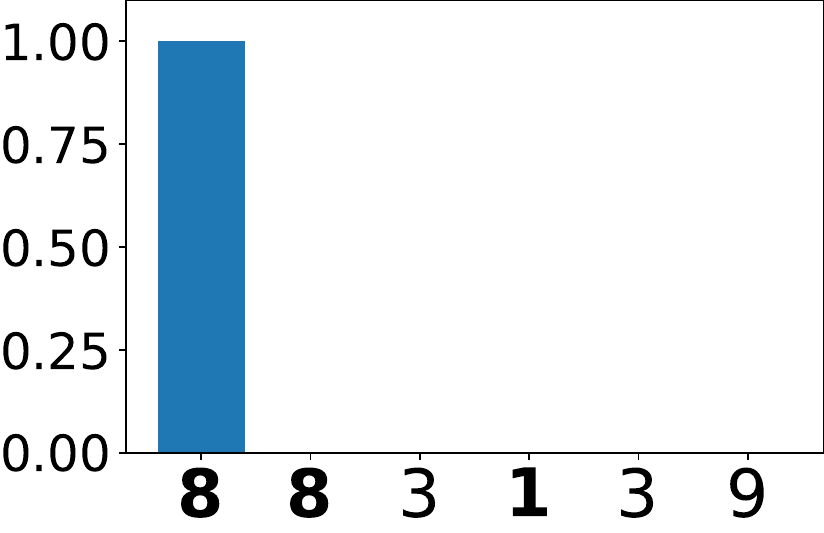}
		\caption{Erasure search.}
		\label{fig:toy-erasure}
	\end{subfigure}
	~
	\begin{subfigure}[b]{0.23\textwidth}
		\centering
		\includegraphics[scale=.2]{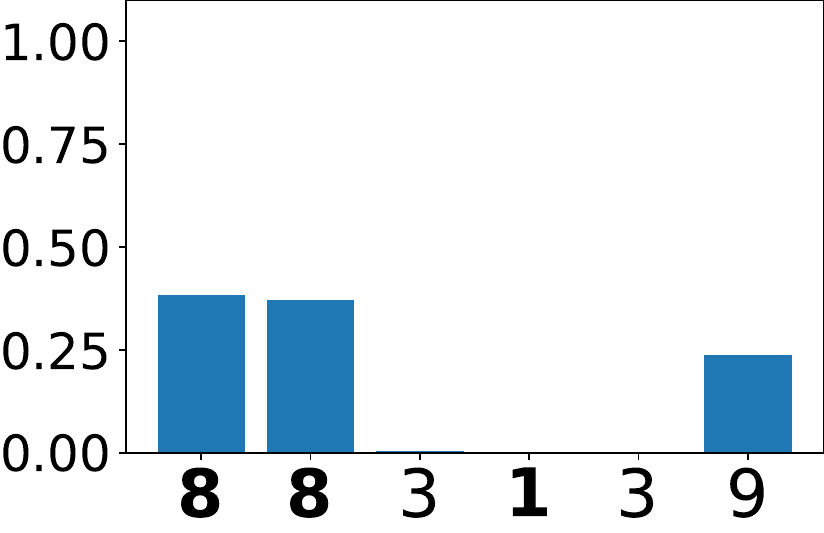}
		\caption{\citet{schulz2020restricting}.}
		\label{fig:toy-schulz}
	\end{subfigure}
    \par\vspace{6pt}
	\begin{subfigure}[b]{0.23\textwidth}
		\centering
		\includegraphics[scale=.2]{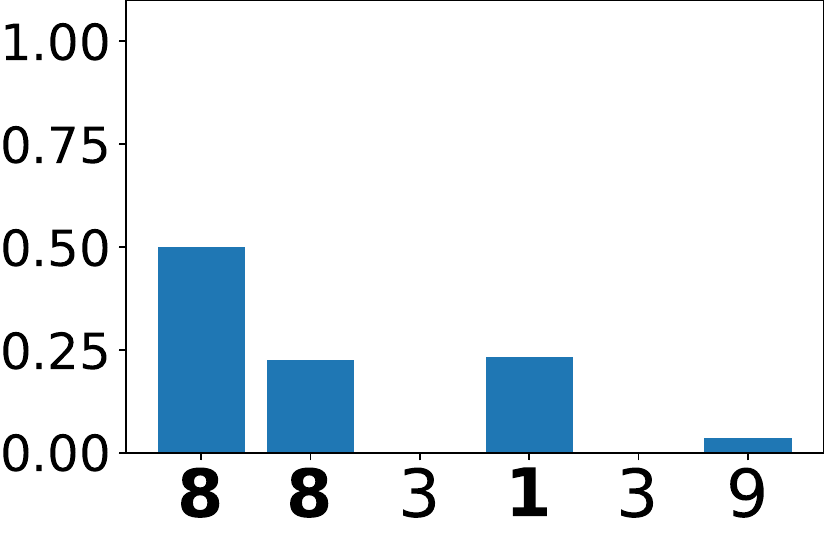}
		\caption{\citet{sundararajan2017axiomatic}.}
		\label{fig:toy-ig}
	\end{subfigure}
    ~
	\begin{subfigure}[b]{0.23\textwidth}
		\centering
		\includegraphics[scale=.2]{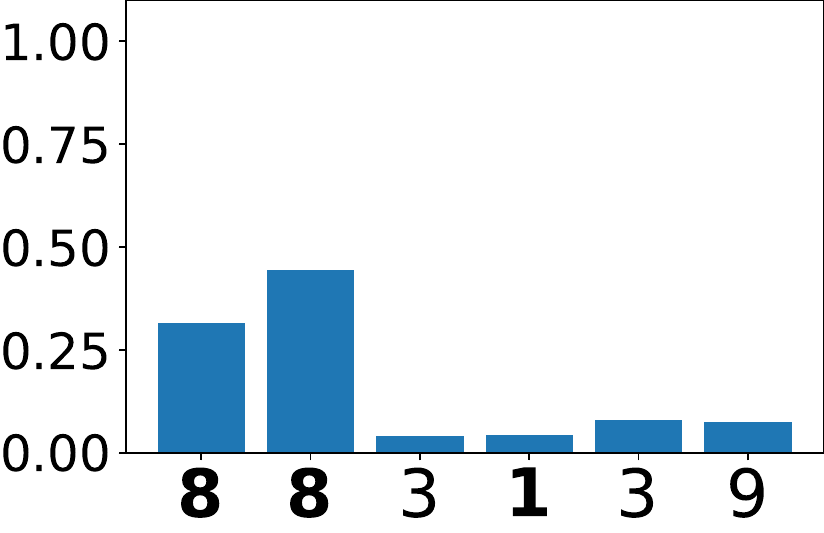}
		\caption{\citet{guan2019towards}}
		\label{fig:toy-guan}
	\end{subfigure}
	\par\vspace{6pt}
	\begin{subfigure}[b]{0.4\textwidth}
		\centering
		\includegraphics[scale=.2]{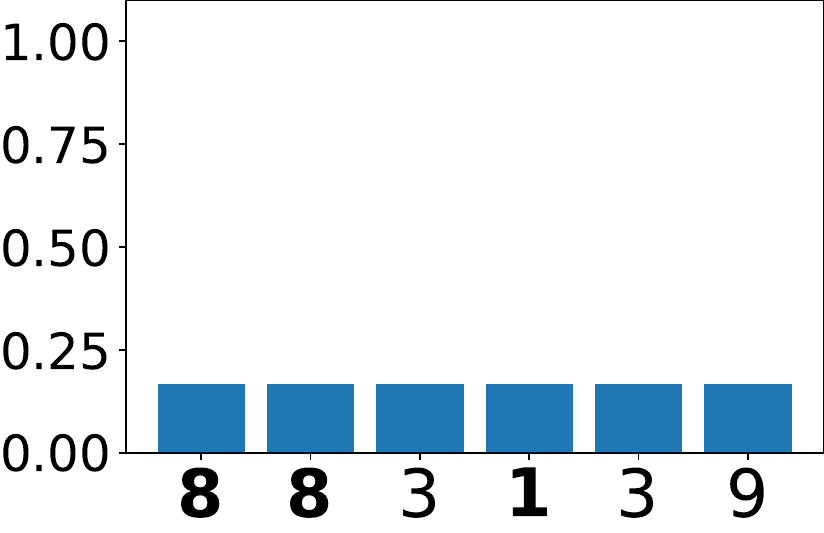}
		\hspace{6pt}
		\includegraphics[scale=.2]{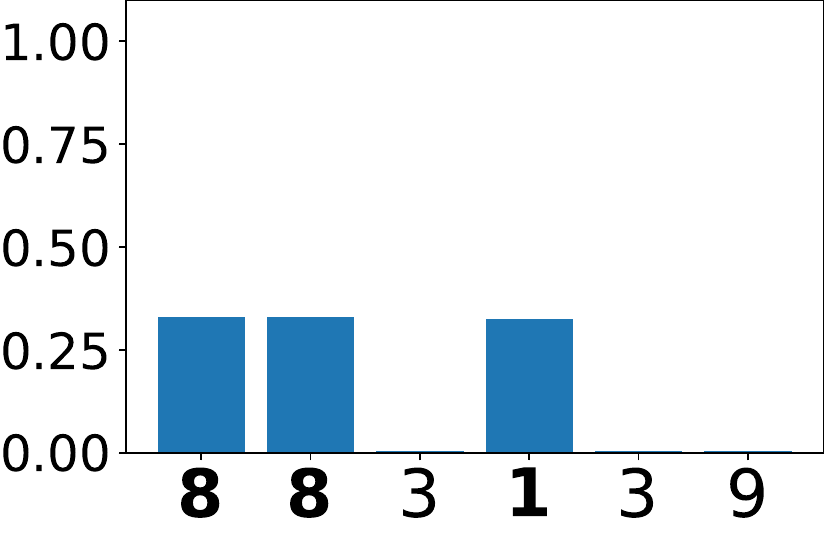}
		\caption{Our \textsc{DiffMask} conditioned on embedding layer (left) and hidden states (right).}
		\label{fig:toy-ours}
	\end{subfigure}
	\caption{Input attributions of several methods on a toy task: Given a sequence $x$ of digits and a query $\langle n, m \rangle$ ($8$ and $1$ in this example) of two digits, determine whether there are more $n$ than $m$ in $x$. Attributions are computed at the vector level and normalized to sum to $1$.}
	\label{fig:toy}
\end{figure}

Despite its conceptual simplicity, subset erasure is not commonly used in practice. First, it is generally {\bf intractable}, and beam search~\cite{feng2018pathologies} or leave-one-out estimates~\cite{zintgraf2017visualizing} are typically used instead. These approximations may be inaccurate. For example, leave-one-out can underestimate the contribution of features due to saturation~\cite{shrikumar2017learning}. More importantly, even these approximations remain very expensive with modern deep (e.g., BERT-based;~\citealp{devlin2018bert}) models, as they require multiple computation passes through the model.  Second, the method is {\bf susceptible to the hindsight bias}: the fact that a feature {\it can be} dropped does not mean that the model `knows' that it can be dropped and that the feature {\it is not} used by the model when processing the example. This results in over-aggressive pruning that does not reflect what information the model uses to arrive at the decision. The issue is pronounced in NLP tasks~(see Figure~\ref{fig:squad-comparison-exact} and~\citealp{feng2018pathologies}), though it is easier to see on an artificial example (Figure~\ref{fig:toy-erasure}). A model is asked to predict if there are more $8$s than $1$s in the sequence. The erasure attributes the prediction to a single $8$ digit, as this reduced example yields the same decision as the original one. However, this does not reveal what the model was relying on: it has counted digits $8$ and $1$ as otherwise, it would not have achieved the perfect score on the test set.

We propose a new method, Differentiable Masking (\textsc{DiffMask}), which overcomes the aforementioned limitations and results in attributions that are more informative and help us understand how the model arrives at the prediction. \textsc{DiffMask} relies on learning sparse stochastic gates (a.k.a., masks), guaranteeing that the information from the masked-out inputs does not get propagated while maintaining end-to-end differentiability without having to resort to REINFORCE~\cite{williams1992simple}. 
The decision to include or disregard an input token is made with a simple model based on intermediate hidden layers of the analyzed model (see Figure~\ref{fig:model-input}). First, this {\it amortization} circumvents the need for combinatorial search making the approach efficient at test time. Second, as with probing classifiers~\cite{adi2017fine,belinkov2019analysis}, this reveals whether the network `knows' at the corresponding layer what input tokens can be disregarded. 
During training inputs are \emph{truly} masked whenever we sample zeros. After training, attribution scores correspond to the expectation of sampling non-zeros.

The amortization lets us not only plot attribution heatmaps, as in Figure~\ref{fig:squad-comparison-ours}, but also analyze how decisions are formed across network layers. In our artificial example, we see that in the bottom embedding layer the model cannot discard any tokens, as it does not `know' which digits need to be counted (Figure~\ref{fig:toy-ours}, left). In the second layer, it `knows' that these are $8$s and $1$s, so the rest gets discarded (Figure~\ref{fig:toy-ours}, right). In question answering (see Figure~\ref{fig:squad-visual-input}), where we use a $24$-layer model,  it takes $13$--$16$ layers for the model to `realize' that {\it `Santa Clara Marriott'} is not relevant to the question and discard it. We also adapt our method to measuring the importance of intermediate states rather than inputs. This, as we discuss later, lets us analyze which states in every layer store information crucial for making predictions, giving us insights about the information flow.

\paragraph{Contributions} 
We introduce \textsc{DiffMask}, a technique addressing limitations of attribution-based methods (especially erasure and its approximations), and demonstrate that it is stable and faithful to the analyzed models. We then use this technique to analyze BERT-based models fined-tuned on sentiment classification and question answering. 

\section{Method} \label{sec:method}
We aim to understand how a trained model processes an input (i.e., a sequence of embedded tokens) to produce an output (e.g., a vector of class probabilities). First, for an input $x = \langle x_1, \ldots, x_n \rangle$, we obtain the output $y=f(x)$ of the model along with its hidden states $\langle h^{(0)}, \ldots, h^{(L)} \rangle$, where $h^{(0)}=x$. We then probe the model using a shallow \emph{interpreter} network which takes hidden states up to a certain layer $\ell$ and outputs a binary mask $z = \langle z_1, \ldots, z_n\rangle$ indicating which input tokens are necessary and which can be disregarded. To assess whether the masked input $\hat x = \langle \hat x_1, \ldots, \hat x_n \rangle$ is sufficient, we re-feed the model with it and compute the output $\hat y = f(\hat x)$. As long as $\hat y$ approximates the original output $y$ well, we deem the inputs masked by $z$ unnecessary.

Masking, however, as in multiplication by \emph{zero}, makes a strong assumption about the geometry of the feature space, in particular, it assumes that the zero vector bears no information. Instead, we replace some of the inputs by a learned baseline vector $b$, i.e., $\hat x_i = z_i \cdot x_i  + (1 - z_i) \cdot b$.

See Figure~\ref{fig:model-input} for an overview. The interpreter model consists of $L+1$ classifiers, the $\ell$th of which conditions on the stack of hidden states up to $h^{(\ell)}$ to predict binary `votes' $v^{(\ell)} = g_\phi^{(\ell)}(h^{(0)}, \ldots, h^{(\ell)})$ towards keeping or masking input tokens. Each classifier is a one-hidden-layer MLP, details and hyperparameters are provided in Appendix~\ref{sec:architecture}. For a given depth $\ell$, the interpreter decides to mask $x_i$ out as soon as $v_i^{(k)} = 0$ for some $k \le \ell$, i.e., $z_i = \prod_{k=0}^{\ell} v_i^{(k)}$. That is, in order to deem $x_i$ unnecessary, it is sufficient to do so based on any subset of hidden states up until $h^{(\ell)}$.

Clearly, there is no direct supervision to estimate the parameters $\phi$ of the probe and the baseline $b$, thus we borrow erasure's objective: namely, we train the probe to mask-out as many input tokens as possible constrained to keeping $f(\hat x) \approx f(x)$. Since often, the output of $f$ parameterizes a likelihood (e.g., a categorical distribution), we formulate the constraint in terms of a divergence $\mathrm{D_\star}$ between the two functions' outputs. We cast this, rather naturally, in the language of constrained optimization.

\paragraph{Objective}
A practical way to minimize the number of non-zeros predicted by $g$ is minimizing the $L_0$ `norm'.\footnote{$L_0$, denoted $\|z \|_0$ and defined as $\#(i|z_i \neq 0)$, is the number of non-zeros entries in a vector. Contrary to $L_1$ or $L_2$, $L_0$ is not a homogeneous function and, thus, not a proper norm. However, contemporary literature refers to it as a norm, and we do so as well to avoid confusion.} Thus, our $\gL_0$ loss is defined as the total number of positions that are not masked:
\begin{equation} \label{eq:l0}
	\gL_0 (\phi,b|x) = \sum_{i=1}^n \mathbf{1}_{[\R_{\neq 0}]}(z_i) \;,
\end{equation}
where $\mathbf{1}(\cdot)$ is the indicator function. We minimize $\gL_0$ for all data-points in the dataset $\gD$ subject to a constraint that predictions from masked inputs have to be similar to the original model predictions:
\begin{equation}  \label{eq:constrained_objective}
	\begin{aligned}
		\min_{\phi,b} & \quad \sum_{x \in \gD} \gL_0 (\phi,b|x) \\
		\mathrm{s.t.} & \quad \mathrm{D_\star}[y \| \hat y] \le m \quad \forall x \in \gD \;, 
	\end{aligned}
\end{equation}
where $\hat y = f(\hat x)$, $y = f(x)$, and the margin $m \in \R_{>0}$ is a hyperparameter. Since non-linear constrained optimisation is generally intractable, we employ Lagrangian relaxation~\citep{boyd2004convex} optimizing instead
\begin{equation} \label{eq:lagrangian_objective}
	\max_\lambda \min_{\phi,b} \sum_{x \in \gD} \gL_0 (\phi,b|x) + \lambda( \mathrm{D_\star}[y \| \hat y] - m) \;,
\end{equation}
where $\lambda \in \R_{\geq 0}$ is the Lagrangian multiplier. 

\paragraph{Stochastic masks}
Our objective poses two challenges: i) $L_0$ is discontinuous and has zero derivative almost everywhere, and ii) to output binary masks, $g$ needs a discontinuous output activation such as the step function. A strategy to overcome both problems is to make the binary variables stochastic and treat the objective in expectation, in which case
one option is to resort to REINFORCE~\citep{williams1992simple}, another is to use a sparse relaxation to binary variables~\cite{louizos2017learning,bastings2019interpretable}. As we shall see (we compare the two aforementioned options in Table~\ref{tab:sst-exact_comparison} and discuss them in Section~\ref{sec:sst}), the latter proved more effective. Thus we opt to use the Hard Concrete distribution, a mixed discrete-continuous distribution on the closed interval $[0, 1]$. This distribution assigns a non-zero probability to exactly zero while it also admits continuous outcomes in the unit interval via the \textit{reparameterization trick}~\citep{kingma2013auto}. We refer to~\citet{louizos2017learning} for details, but also provide a brief summary in Appendix~\ref{sec:binary_concrete}. With stochastic masks, the objective is computed in expectation, which addresses both sources of non-differentiability. Note that during training inputs are \emph{truly} masked-out whenever we sample exact zeros. After training, attribution scores correspond to the expectation of sampling non-zero masks since any non-zero value corresponds to a leak of information.

\paragraph{Masking hidden states} \label{sec:method-hidden}
To reveal which hidden states store information necessary for realizing the prediction, we modify the probe slightly. For a given depth $\ell$, we use a mask $z^{(\ell)} = g_\phi^{(\ell)}(h^{(\ell)})$ to replace some of the \emph{states} in $h^{(\ell)} = \langle h^{(\ell)}_1, \ldots, h^{(\ell)}_n \rangle$ by a layer-specific baseline $b^{(\ell)}$, i.e. $\hat h_i^{(\ell)} = z_i^{(\ell)} \cdot h_i^{(\ell)} + (1 - z^{(\ell)}) \cdot b^{(\ell)}$. The resulting state $\hat h^{(\ell)}$ is used to re-compute subsequent states, $\hat h^{(\ell+1)}, \ldots, \hat h^{(L)}$, as well as the output, which we denote by $\hat y$. Here we do not aggregate `votes' with a product because for this probe we want to discover whether hidden states are predictive of their own \textit{usefulness}. See Figure~\ref{fig:model-hidden} in Appendix~\ref{sec:extra} for an overview of this variant of \textsc{DiffMask}.

\section{Experiments} \label{sec:experiments}
The goal of this work is to uncover a {\it faithful} interpretation of an existing model, i.e. revealing, as accurately as possible, the process by which the model arrives at the prediction. Human-provided labels, such as human rationales~\cite{camburu2018snli,deyoung2019eraser}, will not help us in demonstrating this, as humans cannot judge if an interpretation is faithful~\cite{jacovi2020towards}. More precisely, human-provided labels do not show how the model behaves -- e.g., annotations of what parts of the input are relevant for solving a particular task do \textit{not} constitute a guarantee that a model relies on those parts more than others when making a prediction. When we evaluate an attribution method by comparing its outputs with human annotations, we are not measuring whether it provides faithful attributions but only if they are \textit{plausible} according to humans. This goes against our goals as we aim to use the interpretation method to detect model deficiencies, which are usually cases where the model does not behave like humans. The ground-truth explanations of how a model makes certain predictions depend not only on the data but also on the model, and, unfortunately, are generally not known for real tasks and with complex models. This makes the evaluation and comparison of attribution methods non-trivial.

Our strategy is to i) show the effectiveness of \textsc{DiffMask} in a controlled setting (i.e., a toy task) where ground-truth is available; ii) test the effectiveness of our relaxation for learning discrete masks (on a real model for sentiment classification); and iii) demonstrate that the method is stable and models behave the same when masking is applied. Once we have established that \textsc{DiffMask} can be trusted, we use it to analyze BERT-based models~\citep{devlin2018bert} fine-tuned on sentiment classification, and on question answering. We report hyperparameters in Appendix~\ref{sec:hyper}, and additional plots, examples and analysis in Appendix~\ref{sec:extra}.

\subsection{Toy task} \label{sec:toy}
Our toy task is defined as: given a sequence $x$ of digits (i.e., 
$x_i \in \{0,\cdots,9\}$), and a query $\langle n, m \rangle$ of two digits, determine whether $\#n\!>\!\#m$ in $x$.

\paragraph{Model}
The query and input are embedded, concatenated, and then fed to a single-layer feed-forward NN, followed by a single-layer unidirectional GRU~\citep{cho2014learning}.\footnote{We use a feed-forward NN to incorporate the query information, rather than another GRU layer, to ensure that counting cannot happen in the first layer. This helps us define the ground-truth for the method.} The classification is done by a linear layer that acts on the last hidden state of the GRU. See Appendix~\ref{sec:hyper-toy} for all hyperparameters and a more precise definition of the architecture. Unsurprisingly, the model solves the task almost perfectly (accuracy on test is $>\!99\%$).  

\paragraph{Ground-truth for hidden-state attribution} 
We plot the distribution of hidden states (we use dimensionality $2$, with the purpose of having a bottleneck and to support clear visualization) and observe a linear separation between states of digits present in the query and states not in the query. This means that the role of the feed-forward layer is to decide which digits to keep. Since the model solves the task, the role of the GRU must then be to count which digit occurred the most. The prediction must be attributed uniformly to \emph{all} the hidden states corresponding to either $n$ or $m$. For completeness, Figure~\ref{fig:toy-hidden-projection} in the Appendix~\ref{sec:toy-extra} shows this plot.

\begin{table}[t]
	\centering
    \begin{tabular}{lr}
    \toprule
    \textbf{Methods} &  $\mathrm{D_{JS}} \downarrow$ \\
    \midrule
    Exact erasure                     &   $0.27$ \\
    Integrated Gradient &  $0.27$ \\
    \citet{schulz2020restricting}     &   $0.18$ \\
    \citet{guan2019towards}           &   $0.24$ \\
    \textsc{DiffMask}                 &  $\mathbf{0.00}$ \\
    \bottomrule
    \end{tabular}
	\caption{Toy task: attribution to hidden states, average divergence in nats between the \textit{ground-truth} attributions and those by different methods. *The Delta distribution does not share support with the \textit{ground-truth}.}
	\label{tab:toy-metrics}
\end{table}

\paragraph{Results}
We start with an example of \emph{input attributions}, see Figure~\ref{fig:toy}, which illustrates how \textsc{DiffMask} goes beyond input attribution as typically known.\footnote{To enable comparison across methods, the attributions in this Section are normalized between $0$ and $1$.} The attribution provided by erasure (Figure~\ref{fig:toy-erasure}) is not informative: for each datapoint the search always finds a single digit that is sufficient to maintain the original prediction and discards all the other inputs. The perturbation methods by~\citet{schulz2020restricting} and~\citet{guan2019towards} (Figure~\ref{fig:toy-schulz} and~\ref{fig:toy-guan}) are also over-aggressive in pruning. They assign low attribution to some items in the query even though those had to be considered when making the prediction. Differently from other methods, \textsc{DiffMask} reveals input attributions conditioned on different levels of depth. Figure~\ref{fig:toy-ours} shows both input attributions according to the input itself and according to the hidden layer. It reveals that at the embedding layer there is no information regarding what part of the input can be erased: attribution is uniform over the input sequence. After the model has observed the query, hidden states predict that masking input digits other than $n$ and $m$ will not affect the final prediction: attribution is uniform over digits in the query. This reveals the role of the feed-forward layer as a filter for positions relevant to the query. Other methods do not allow for this type of inspection. These observations are  consistent across the entire test set. 

For \emph{attribution to hidden states} (i.e., the output of the feed-forward layer) we can compare methods in terms of how much their attributions resemble the ground-truth across the test set. Table~\ref{tab:toy-metrics} shows how the different approaches deviate from the gold-truth in terms of Kullback-Leibler ($\mathrm{D_{KL}}$) and Jensen–Shannon ($\mathrm{D_{JS}}$) divergences.\footnote{We use $\mathrm{D_{KL}}[p\|q]$ and $\mathrm{D_{JS}}[p\|q]$ where $p$ is the \textit{ground-truth} distribution and $q$ is the predicted attribution distribution.}

\subsection{Sentiment Classification} \label{sec:sst}
We turn now to a real task and analyze models fine-tuned for sentiment classification on the Stanford Sentiment Treebank ~\citep[SST;][]{socher2013recursive}. 

\paragraph{Erasure search as learning masks} \label{sec:sst-erasure-quality}
Before diving into an analysis of a BERT sentiment model, we would like to demonstrate that we can approximate the result of erasure well through our differentiable relaxations. For that, we train a single-layer GRU sentiment classifier and compare the analyses by \textsc{DiffMask} to solutions provided by erasure (exact search). To isolate the impact of our objective, we disable amortization, thus estimating Hard Concrete parameters for each example independently. We compare \textsc{DiffMask} to REINFORCE~\citep{williams1992simple} with a moving average baseline for variance reduction. Since erasure is prohibitive for long sentences, we limit our evaluation to sentences up to $25$ words ($54\%$ of the data). Table~\ref{tab:sst-exact_comparison} shows that \textsc{DiffMask} and REINFORCE achieve comparable levels of sparsity, but our method reaches an optimal solution much more often (33\% of the times vs 9\%) and is, on average, closer to an optimal solution (81\% $\mathrm{F_1}$ vs 75\% $\mathrm{F_1}$).

\begin{table}[t]
	\centering
    \begin{tabular}{lrr}
    \toprule
    \textbf{Metric}  & REINFORCE+          & \textsc{DiffMask} \\
    \midrule
    Precision        &             $74.69$ &           $\mathbf{81.26}$ \\
    Recall           &             $80.82$ &           $\mathbf{85.89}$ \\
    $\mathrm{F}_1$   &             $73.57$ &           $\mathbf{80.75}$ \\
    Optimality       &              $8.83$ &           $\mathbf{32.67}$ \\
    $L_0$            &             $33.13$ &           $\mathbf{30.58}$ \\
    \bottomrule
    \end{tabular}
	\caption{Sentiment classification: optimization with \textsc{DiffMask} and REINFORCE (not amortised -- with a moving average baseline for variance reduction) vs. erasure with exact search. All metrics are computed at token level; optimality is measured at sentence level.}
	\label{tab:sst-exact_comparison}
\end{table}

\paragraph{Faithfulness and Plausibility} \label{sec:sst-faithfulness-plausibility}
Now, we get back to the fully-amortized \textsc{DiffMask} approach applied to a 12-layers BERT\textsubscript{BASE} model and verify that there is no performance degradation when applying masking. Training hyperparameters are reported in Appendix~\ref{sec:hyper-sst}. The $\mathrm{F_1}$ score of the model on the validation set moved from $37.9\%$ to $38.3\%$ while masking $46.3\%$ input tokens, and to $38.9\%$ while masking $67.6\%$ hidden states. The explanations provided by \textsc{DiffMask} are also stable. Across $5$ runs with different seeds, the standard deviation of input attributions are $0.05$ and $0.03$ for inputs and hidden states, respectively.

\begin{figure}[t]
	\centering
	\begin{subfigure}[b]{0.23\textwidth}
		\centering
	    \includegraphics[width=\textwidth]{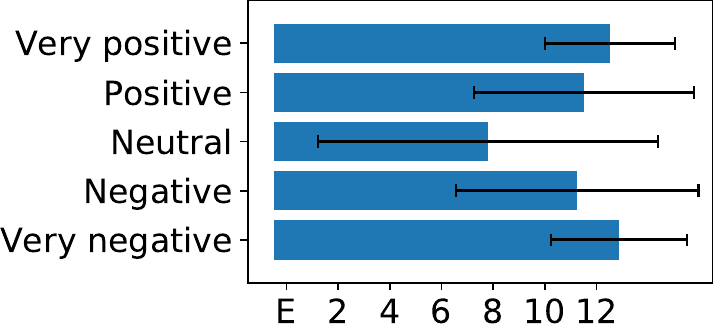}
		\caption{Input.}
	    \label{fig:sst-stats-sentiment-input}
	\end{subfigure}
	~
	\begin{subfigure}[b]{0.23\textwidth}
		\centering
	    \includegraphics[width=\textwidth]{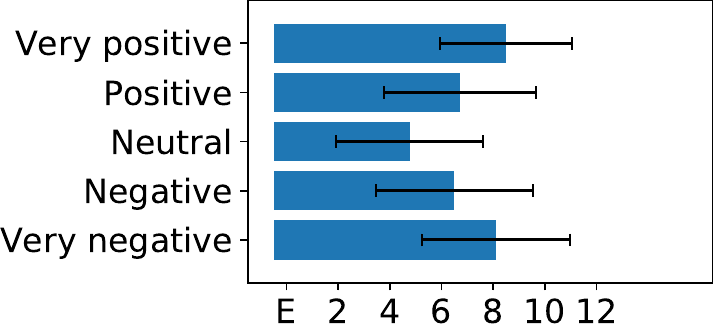}
		\caption{Hidden states.}
	    \label{fig:sst-stats-sentiment-hidden}
	\end{subfigure}
	\caption{Sentiment classification: average number of layers that predict to keep input tokens or hidden states aggregated by token level sentiment annotations.}
	\label{fig:sst-stats-sentiment}
\end{figure}

While we cannot use human labels to evaluate faithfulness of our method, comparing them and \textsc{DiffMask} attribution will tell us whether the sentiment model relies on the same cues as humans. Specifically, we compare to SST token level annotation of sentiment. In Figure~\ref{fig:sst-stats-sentiment-input}, we show after how many layers on average an input token is dropped, depending on its sentiment label. This suggests that the model relies more heavily on strongly positive or negative words and, thus, is generally consistent with human judgments (i.e., \textit{plausible}).

\begin{figure}[t]
	\centering
	\begin{subfigure}[b]{0.23\textwidth}
		\centering
		\includegraphics[width=\textwidth]{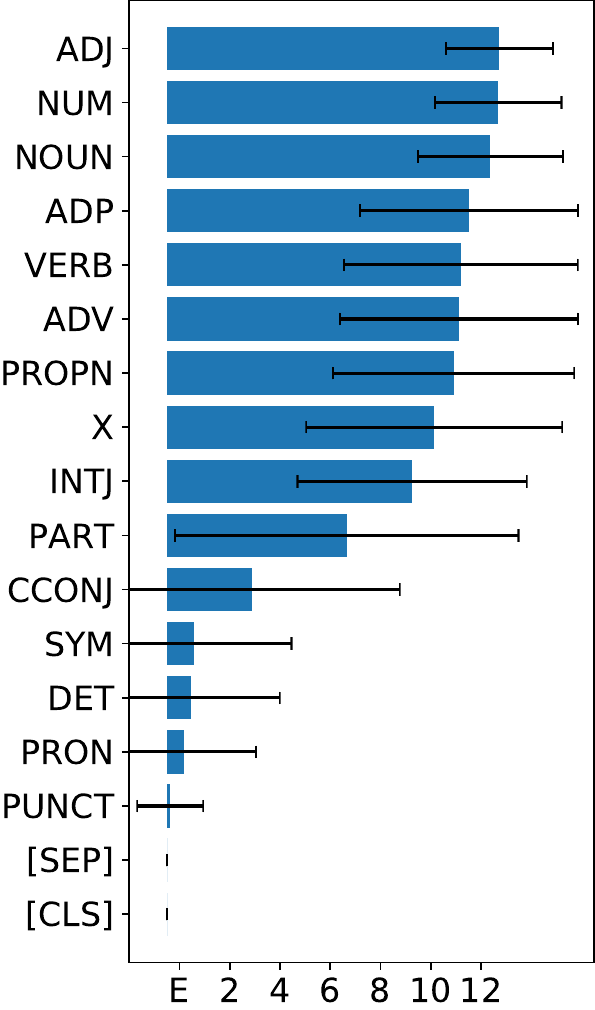}
		\caption{Inputs.}
		\label{fig:sst-stats-pos-input}
	\end{subfigure}
	~
	\begin{subfigure}[b]{0.23\textwidth}
		\centering
		\includegraphics[width=\textwidth]{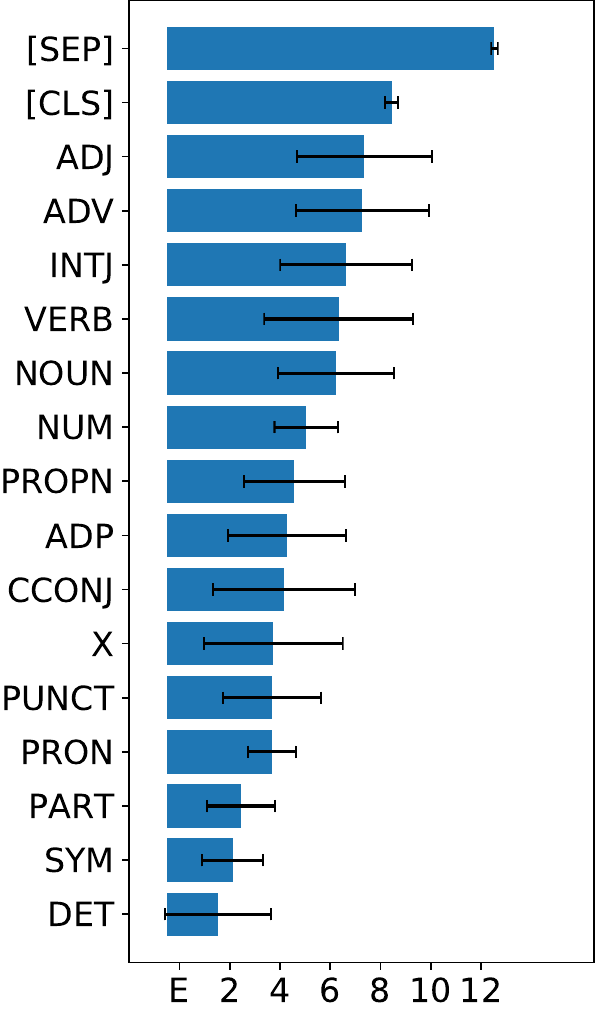}
		\caption{Hidden states.}
		\label{fig:sst-stats-pos-hidden}
	\end{subfigure}
	\caption{Sentiment classification: average number of layers that predict to keep input tokens (a) or hidden states (b) aggregating by part-of-speech tags (POS) and \texttt{[CLS]}, \texttt{[SEP]} tokens on validation set.}
	\label{fig:sst-stat-pos}
\end{figure}

\paragraph{Analysis} \label{sec:sst-analysis}
We used \textsc{DiffMask} to analyse the behavior of our BERT model. In Figure~\ref{fig:sst-stat-pos}, we report the average number of layers that input tokens or hidden states are kept for (or, equivalently, after how many layers they are dropped on average), aggregating by part-of-speech tags (PoS). It turns out that determinants, punctuation, and pronouns can be completely discarded from the input across all validation set, while adjectives and nouns should be kept. Also the $\texttt{[CLS]}$ and $\texttt{[SEP]}$ tokens can be ignored indicating that the model does not need such markers. Examining the POS tags distribution for hidden states leads to further conclusions. Here, the $\texttt{[CLS]}$ and $\texttt{[SEP]}$ tokens are the most important ones. This is not surprising as the classifier on top of BERT uses the $\texttt{[CLS]}$ hidden state which gets progressively updated through all layers. Both these special tokens are not important as inputs because BERT can infer these markers in other layers, however,  they are heavily used in the computation.

\begin{figure}[t]
	\centering
	\begin{subfigure}[b]{0.23\textwidth}
		\centering
		\includegraphics[scale=.3]{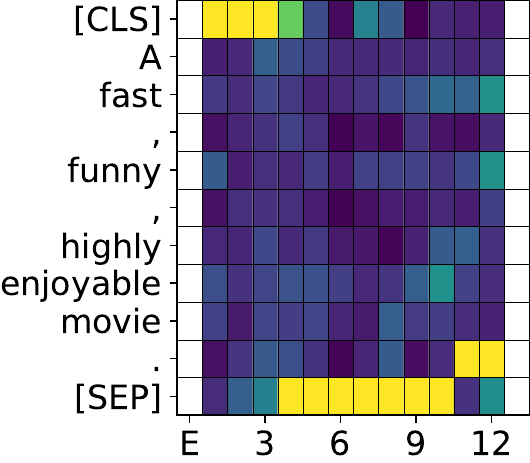}
		\caption{Attention.}
		\label{fig:sst-att}
	\end{subfigure}
	~
	\begin{subfigure}[b]{0.23\textwidth}
		\centering
		\includegraphics[scale=.3]{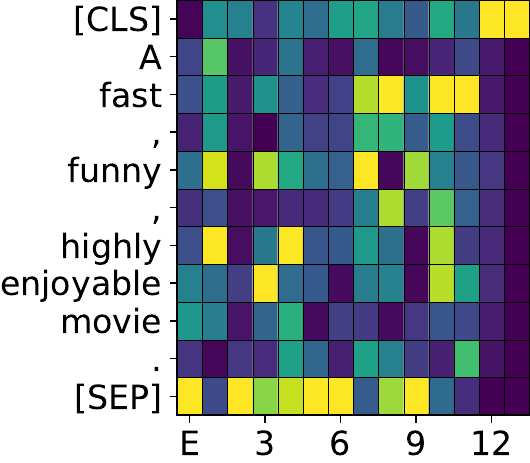}
		\caption{\citet{sundararajan2017axiomatic}.}
		\label{fig:sst-ig}
	\end{subfigure}
	\par\vspace{6pt}
	\begin{subfigure}[b]{0.23\textwidth}
		\centering
		\includegraphics[scale=.3]{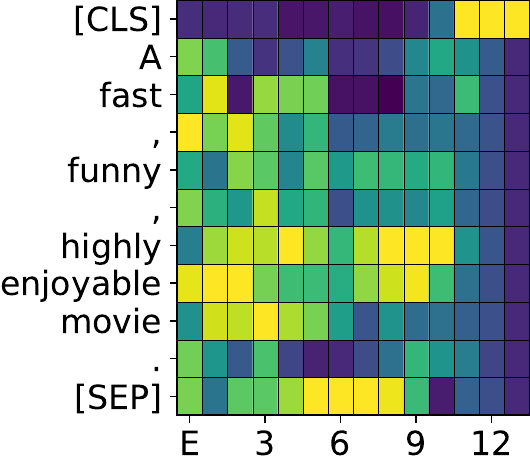}
		\caption{\citet{schulz2020restricting}.}
		\label{fig:sst-schulz}
	\end{subfigure}
	~
	\begin{subfigure}[b]{0.23\textwidth}
		\centering
		\includegraphics[scale=.3]{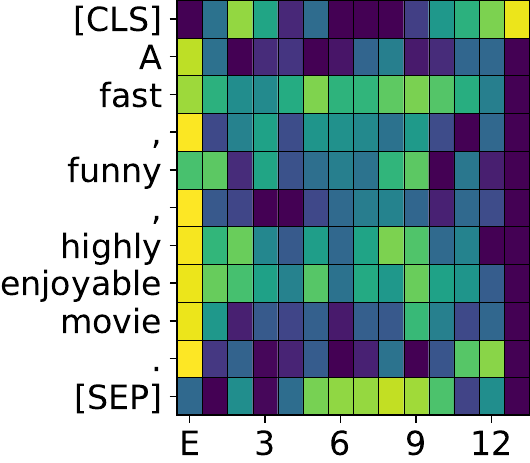}
		\caption{\citet{guan2019towards}.}
		\label{fig:sst-guan}
	\end{subfigure}
	\par\vspace{6pt}
	\begin{subfigure}[b]{0.23\textwidth}
		\centering
		\includegraphics[scale=.3]{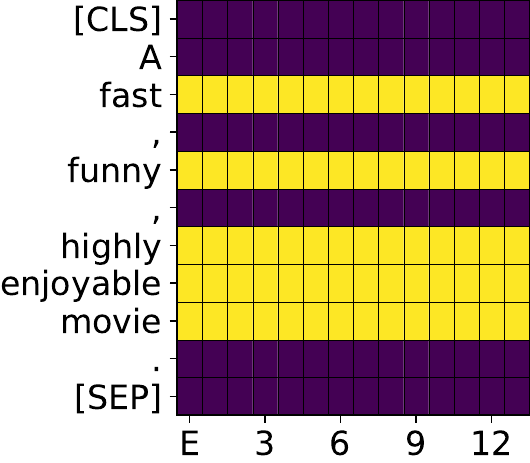}
		\caption{\textsc{DiffMask} on input.}
		\label{fig:sst-ours-input}
	\end{subfigure}
	~
	\begin{subfigure}[b]{0.23\textwidth}
		\centering
		\includegraphics[scale=.3]{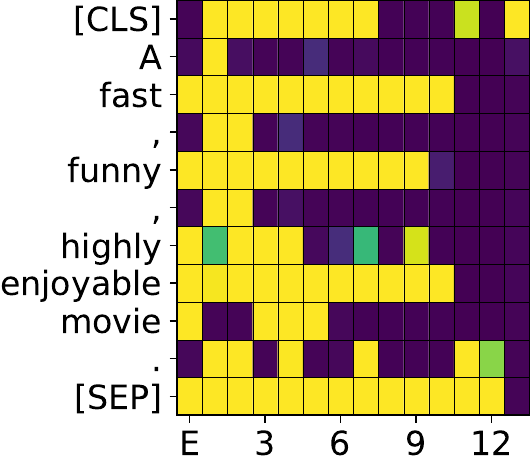}
		\caption{\textsc{DiffMask} on hidden.}
		\label{fig:sst-ours-hidden}
	\end{subfigure}
	\caption{Sentiment classification: comparison between attribution method for hidden layers w.r.t. the predicted label. All plots are normalized per-layer by the largest attribution. Attention heatmap is obtained max pooling over heads and averaging across positions.}
	\label{fig:sst-comparison}
\end{figure}

Figure~\ref{fig:sst-ours-input} we show a visual example of that. We see that the model, even in the bottom layers, knows that the punctuation and both separators can be dropped from the input. This contrasts with hidden states attribution (Figure~\ref{fig:sst-ours-hidden}) which indicates that the separator states (especially $\texttt{[SEP]}$) are very important. By putting this information together, we can hypothesize that the separator is used to aggregate information from the sentence, relying on self-attention. In fact, this aggregation is still happening in layer $12$; at the very top layers, states corresponding to almost all non-separator tokens can be dropped. 

\paragraph{Comparison to other methods} \label{sec:sst-comparisons}
In Figure~\ref{fig:sst-comparison}, we visually compare different techniques on one example form validation set. While previous techniques (e.g., integrated gradient) do not let us test what a model `knows' in a given layer (i.e. attribution to input conditioned on a layer), they can be used to perform attribution to hidden layers. All methods except attention correctly highlight the last hidden state of the $\texttt{[CLS]}$ token as important. Its importance is due to the top-level classifier using the $\texttt{[CLS]}$ hidden state. Although for \textsc{DiffMask} we show the expectation of keeping states, it assigns much sharper attributions. For instance, on the validation set, it assigns to the last hidden state of the $\texttt{[CLS]}$ the biggest attribution $99\%$ of the times where~\citet{schulz2020restricting} only $71\%$. Raw attention (Figure~\ref{fig:sst-att}) does not seem to highlight any significant patterns in that example except that start and end of sentence tokens ($\texttt{[CLS]}$ and $\texttt{[SEP]}$, respectively) receive more attention than the rest.\footnote{\citet{voita2019analyzing} and~\citet{michel2019sixteen} pointed out that many Transformer heads play no or minor role.} Attributions by~\citet{schulz2020restricting} and~\citet{guan2019towards} assign slightly higher importance to hidden states corresponding to `highly' and `enjoyable', whereas it is hard to see any informative patterns provided by integrated gradient. Notice that for \textsc{DiffMask}, a near-zero attribution has a very clear interpretation: such a state is not used for prediction since in expectation it is dropped (not gated).

\subsection{Question Answering} \label{sec:squad}
We turn now to QA where we analyse a fine-tuned BERT\textsubscript{LARGE} model on the Stanford Question Answering Dataset~\citep[\textsc{SQuAD};][]{rajpurkar2016squad}.

\begin{figure}[t]
	\centering
	\begin{subfigure}[b]{0.23\textwidth}
		\centering
	    \includegraphics[scale=.28]{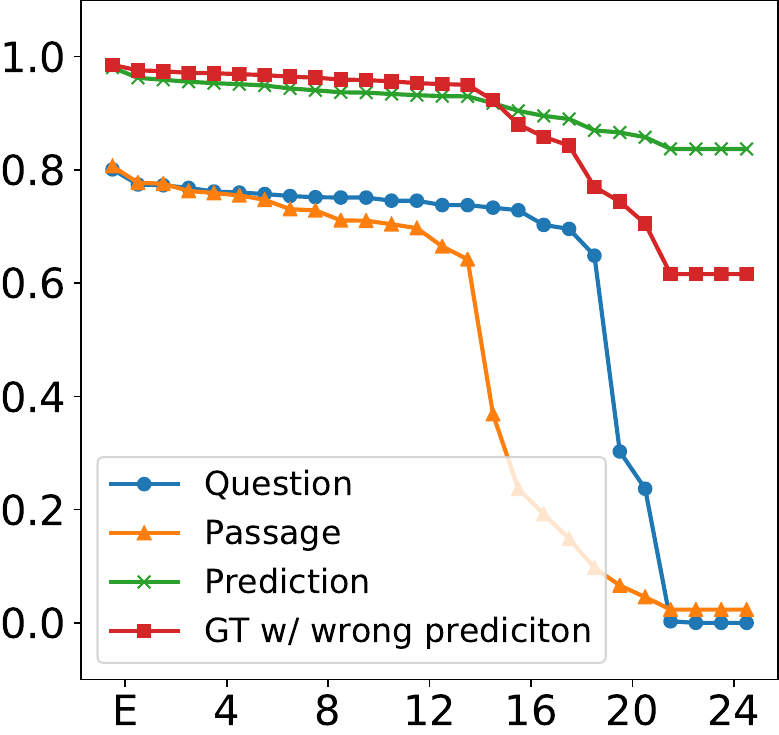}
		\caption{Input.}
	    \label{fig:squad-stats-keep-input}
	\end{subfigure}
	~
	\begin{subfigure}[b]{0.23\textwidth}
		\centering
	    \includegraphics[scale=.28]{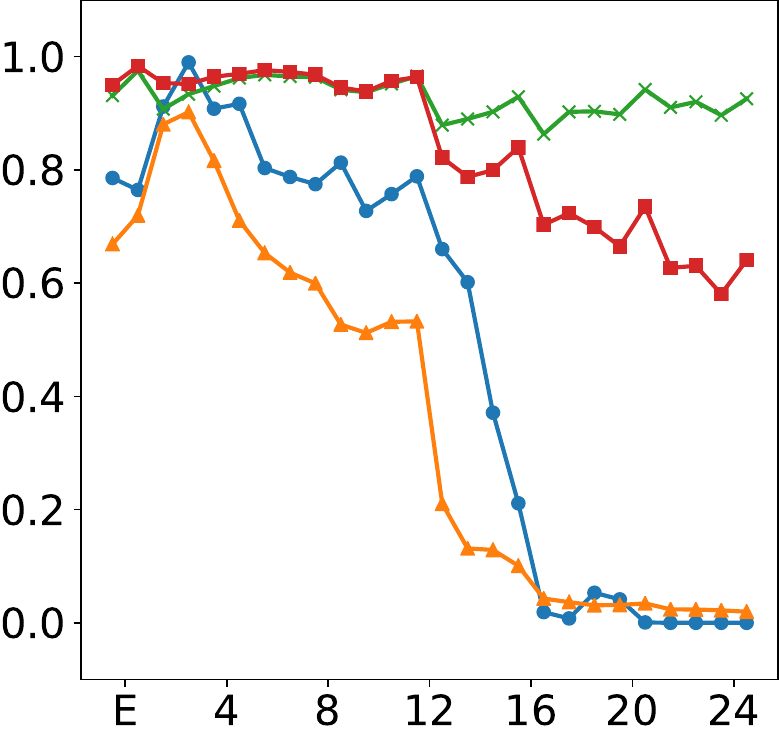}
		\caption{Hidden states.}
	    \label{fig:squad-stats-keep-hidden}
	\end{subfigure}
	\caption{QA: average expectation of keeping input (a) and hidden states (b) from different layers.}
	\label{fig:squad-stats-keep}
\end{figure}

\begin{figure}[t]
    \centering
	\begin{subfigure}[b]{0.23\textwidth}
		\centering
		\includegraphics[scale=0.28]{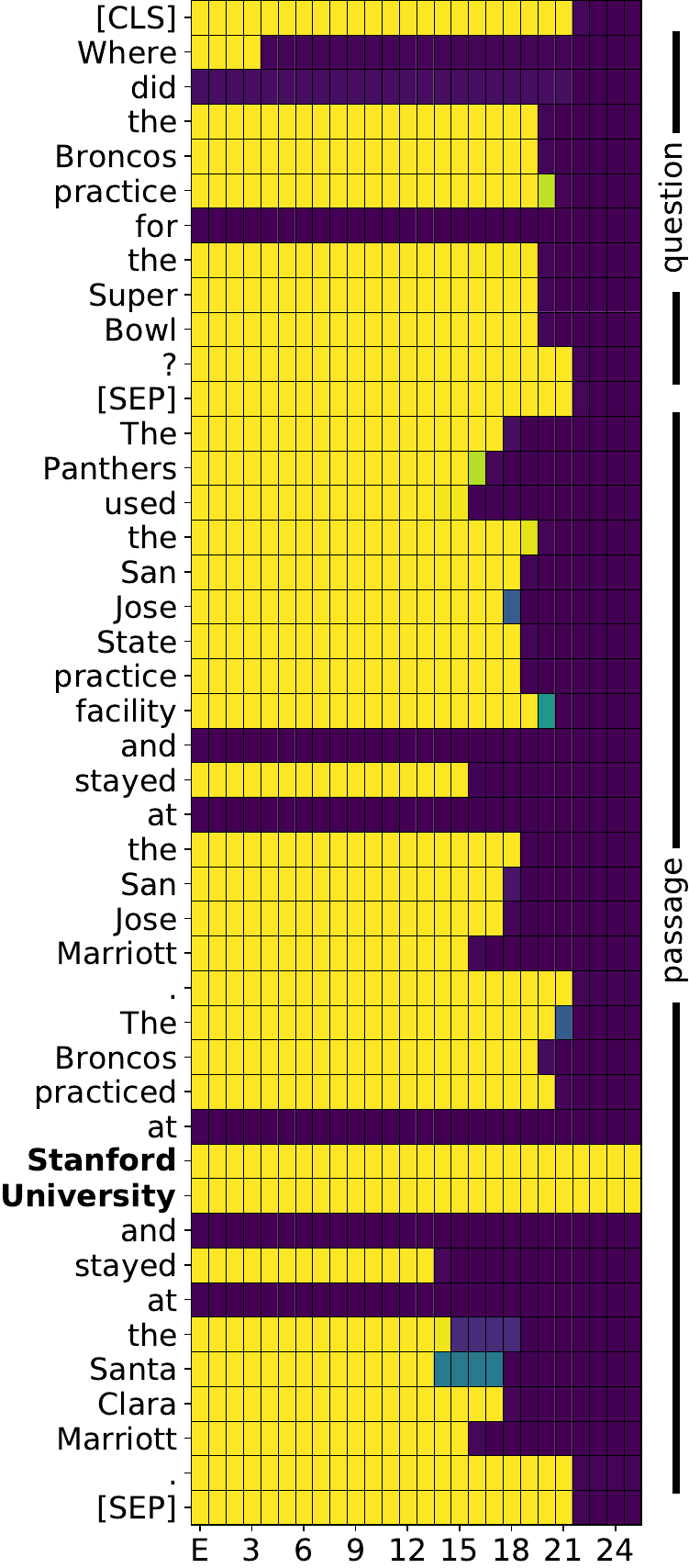}
		\caption{Gating the input.}
		\label{fig:squad-visual-input}
	\end{subfigure}
	~
	\begin{subfigure}[b]{0.23\textwidth}
		\centering
		\includegraphics[scale=0.28]{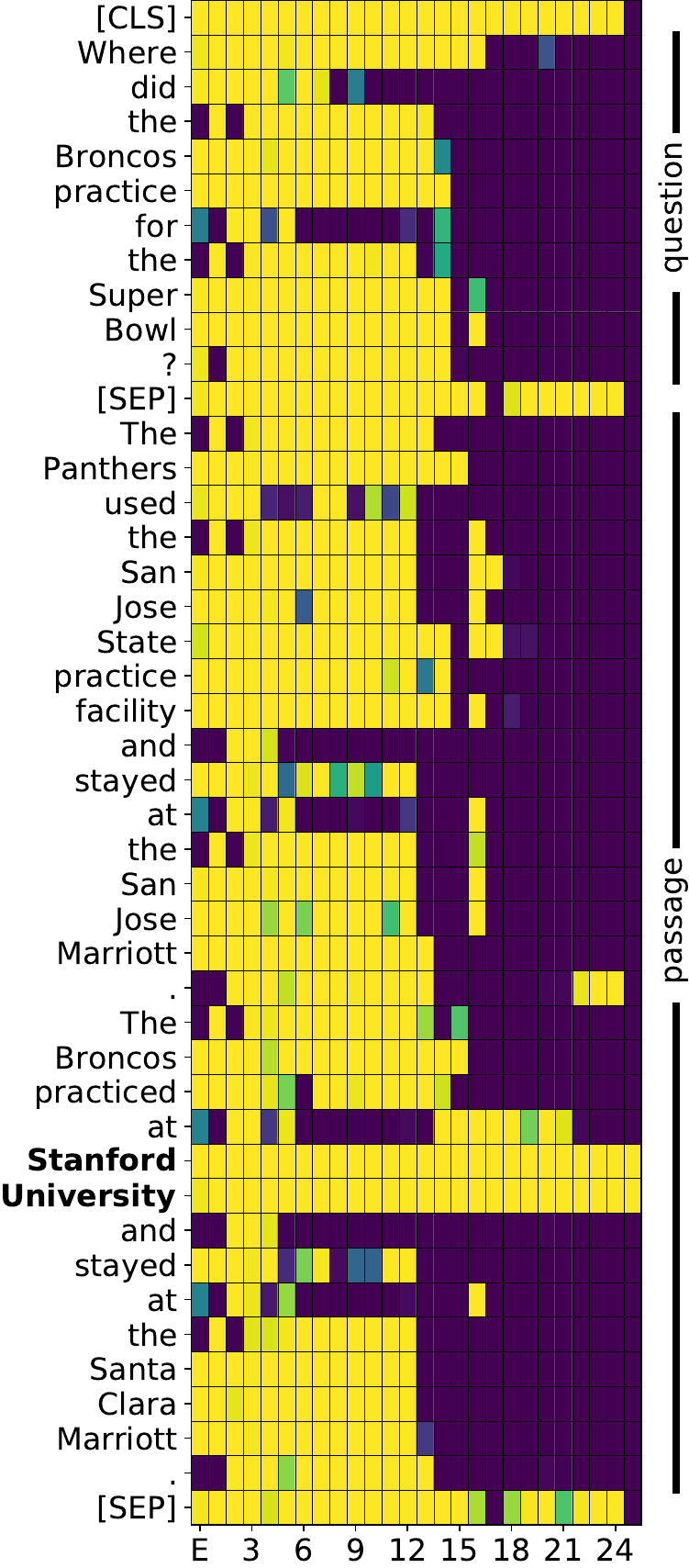}
		\caption{Gating hidden states.}
		\label{fig:squad-visual-hidden}
	\end{subfigure}
	\caption{QA: attribution the inputs (a) and hidden states (b). The correct answers is highlighted in bold.}
	\label{fig:squad-visual}
\end{figure}

\paragraph{Analysis}
We start by asking \textsc{DiffMask} \textbf{which tokens does the model keep?}
We do a similar analysis as for sentiment classification of POS tags over the entire validation set. We summarize the results in Figure~\ref{fig:squad-stat-pos} in Appendix~\ref{sec:sst-extra}. It turns out that conjunctions and adpositions are dropped by the embedding and first layer, respectively, on average. On the contrary, proper nouns and punctuation are usually predicted to be dropped only after the $14$th layer. We argue that due to the pre-training objective, BERT could infer well missing parts of the input, especially if they are trivial to infer (e.g., as often the case for prepositions). On the contrary, nouns and proper nouns are important as they count for $84\%$ of the answers on SQuAD. For example, in Figure~\ref{fig:squad-visual-input}, we can see that it takes $13$--$16$ layers for the model to `realize' that {\it `Santa Clara Marriot'} is not relevant to the question and discard it.

Unlike in sentiment classification, separator tokens as well as punctuation assume a central role as inputs (i.e., punctuation is considered the most important POS tag as for both questions and passages is usually dropped after the $17$th layer). Punctuation serves to demarcate sentence boundaries,  useful for QA but not for sentiment classification.

Tokens from questions are generally masked by higher layers than tokens from passages as we show in Figure~\ref{fig:squad-stats-keep-input}, which suggests that they are more important. We highlight that even in higher layers when \textsc{DiffMask} masks $>\!95\%$ of the tokens, the original model prediction is almost always kept $>\!90\%$. Noticeably, when the original BERT makes wrong predictions, the tokens annotated as the ground truth answer are kept $\sim\!60\%$ of the time. This may suggest that when this happens the model still considers other options (e.g., valid options such as the ground truth) as plausible,  thus \textsc{DiffMask} detects them as important.

Now, we inspect hidden states attributions to answer \textbf{where is the information stored?} In Figure~\ref{fig:squad-stats-keep-hidden} we can see a similar trend as for masking input, i.e., question's hidden states are kept more on average and deeper in the computation. States on layers $2$--$3$ are dropped less than from the embedding and first layer. This is consistent with findings of~\citet{voita2019bottom} which show that frequent tokens, such as determiners, accumulate contextual information. However, they are not important as inputs as we show in an example in Figure~\ref{fig:squad-visual-hidden}.

The hidden states corresponding to separator tokens are always kept across all layers except the last one across the validation set. Notice that, this token is also used as a delimiter between the question and the passage, and hence indicates where questions as well as passages end. 

The level of hidden states pruning is quite incremental (after layer $3$) and gets strong, after layer $9$ more than 50\% of them can be masked out. A steep increase in superfluous states $13$--$14$ (visible on both parts of Figure~\ref{fig:squad-stats-keep}) may indicate that some states, at that point in computation, contain enough information needed for the classification while all the others can indeed be removed without affecting the model prediction. Our observation that higher layers are more predictive is in line with findings of~\citet{kovaleva2019revealing}. They pointed out that the final layers of BERT change most and are more task-specific. Again, the fact that states corresponding to the ground truth answer are still active on top layers when the model makes a wrong prediction indicates that the model is still considering different span options across top layers as well.

\paragraph{Comparison to other methods} \label{sec:squad-comparisons}
As we do not have access to the ground-truth, we start by contrasting \textsc{DiffMask} qualitatively to other attribution methods on a few examples. We highlight some common pitfalls that afflict other methods (such as the hindsight bias) and how \textsc{DiffMask} overcomes those. This helps demonstrate our method's faithfulness to the original model.

Figure~\ref{fig:squad-comparison} shows input attributions by different methods on an example from the validation set. Erasure (Figure~\ref{fig:squad-comparison-exact}), as expected, does not provide useful insights, it essentially singles out the answer discarding everything else including the question. This cannot be faithful and is a simple consequence of erasure's hindsight bias: when only the span that contains the answer is presented as input, the model predicts that very span as the answer, but this does not imply that the model ignores everything else when presented with the complete document as input. The methods of~\citet{schulz2020restricting} and~\citet{guan2019towards} optimize attributions on single examples and thus also converge to assigning high importance mostly to words that support the current prediction and that indicate the question type. For this experiment we used Per-Sample Bottleneck attribution from~\citet{schulz2020restricting}. The authors also proposed a Readout Bottleneck where they train a second neural network to predict the mask. But differently from our formulation, they condition on subsequent layers and thus attributions are prone to the hindsight bias.

Integrated gradient does not seem to highlight any discernible pattern, which we speculate is mainly because a zero baseline is not suitable for word embeddings. Choosing a more adequate baseline is not straightforward and remains an important open issue~\citep{sturmfels2020visualizing}. Note that, \textsc{DiffMask} without amortization (Figure~\ref{fig:squad-comparison-ours-nonamo}) resembles erasure (as shown in \S~\ref{sec:sst-erasure-quality} for SST).

Differently from all other methods, our \textsc{DiffMask} probes the network to understand what it `knows' about the input-output mapping in different layers. In Figure~\ref{fig:squad-comparison-ours} we show the expectation of keeping input tokens conditioned on any one of the layers in the model to make such predictions (see Figure~\ref{fig:squad-visual-input} for a per-layer visualization). Our input attributions highlight that the model, in expectation across layers, \textit{wants} to keep words in the question, the predicate `practice' in both sentences as well as all potential candidate answers (i.e., named entities). But eventually, the most important spans are in the question and the answer itself.

\section{Related Work}
While we motivated our approach through its relation to erasure,
an alternative way of looking at our approach is considering it as a {\it perturbation-based} method. This recently introduced class of attribution methods~\cite{ying2019gnnexplainer, guan2019towards, schulz2020restricting,taghanaki2019infomask}, instead of erasing input, injects noise. Besides back-propagation and attention-based methods discussed in the introduction, another class of interpretation methods~\cite{murdoch2017automatic,singh2018hierarchical,jin2019towards} builds on prior work in cooperative game theory (e.g., Shapley value of~\citealp{shapley1953value}). These methods are not trivial to apply to a new model, as they are architecture-specific. Their hierarchical versions (e.g.,~\citealp{singh2018hierarchical,jin2019towards}) also make a strong assumption about the structure of interaction (e.g., forming a tree) which may affect their faithfulness. Also~\citet{chen2018learning} share some similarities to our work as they also do amortization but use the \textit{Gumbel softmax trick}~\citep{maddison2016concrete,jang2016categorical} to approximate minimal subset selection. They assume that the subset contains exactly $k$ elements where $k$ is a hyperparameter. Moreover, their explainer is a separate model predicting input subsets, rather than a `probe' on top of the model's hidden layers, and hence cannot be used to reveal how decisions are formed across layers.

A large body of literature analyzed BERT and Transformed-based models. For example,~\citet{tenney2019you} and~\citet{van2019does} probed BERT layers for a range of linguistic tasks, while~\citet{hao2019visualizing} analyzed the optimization surface. \citet{rogers2020primer} provides a comprehensive overview of recent BERT analysis papers.

There is a stream of work on learning interpretable models by means of extracting latent rationales~\citep{lei-etal-2016-rationalizing,bastings2019interpretable}. Some of the techniques underlying \textsc{DiffMask} are related to that line of work. They employ stochastic masks to \emph{learn an interpretable model}, which they train by minimizing a downstream loss subject to constraints on $L_0$, whereas we employ stochastic masks to \emph{interpret an existing model}, and for that, we minimize $L_0$ subject to constraints on that model's output distribution. In our very recent work~\citet{schlichtkrull2020interpreting}, we also employ stochastic masks and $L_0$ regularization for analyzing graph neural networks. We learn which edges are relevant in multi-hop question answering and graph-based semantic role labeling~\citep{marcheggiani2017encoding,decao2019question}.

\section{Conclusion}
We have introduced a new \textit{post hoc} interpretation method which learns to completely remove subsets of inputs or hidden states through masking. We circumvent an intractable search by learning an end-to-end differentiable prediction model. To overcome the hindsight bias problem, we probe the model's hidden states at different depths and amortize predictions over the training set. Faithfulness is validated in a controlled experiment pointing more clearly to some flaws of other attribution methods. We used our method to study BERT-based models on sentiment classification and question answering. \textsc{DiffMask} sheds light on what different layers `know' about the input and where information about the prediction is stored in different layers. 

\newpage
\subsection*{Acknowledgements}
The authors want to thank Christos Baziotis, Elena Voita, Dieuwke Hupkes, and Naomi Saphra for helpful discussions. This project is supported by SAP Innovation Center Network, ERC Starting Grant BroadSem (678254), the Dutch Organization for Scientific Research (NWO) VIDI 639.022.518, and the European Union's Horizon 2020 research and innovation programme under grant agreement No 825299 (Gourmet).

\bibliography{emnlp2020}
\bibliographystyle{acl_natbib}

\clearpage
\appendix

\section{Probe parameterization} \label{sec:architecture}
We parameterized the probe functions with a single layer MLP. Note that the architecture of this probe is chosen to be simple but different model choices are also possible and will not affect our general framework.\footnote{In our open source implementation, we also used different architectures. Final results did not change much.} When masking input tokens, `votes' are computed as $v_i^{(\ell)} = g_\phi^{(\ell)}(h_i^{(\ell)})$ where
\begin{align}
    \gamma_i^{(\ell)}  & = \xi \cdot \tanh \left( \operatorname{NN}^{(\ell)} ( [x_i;h_i^{(\ell)}] ) \right) + b^{(\ell)} \;, \\
	v_i^{(\ell)} & \sim \mathrm{Hard Concrete}(v_i^{(\ell)} ; \tau, \gamma_i^{(\ell)},  l, r) \;,
\end{align}
where $\xi=10, \tau = 0.2, l=-0.2, r=1.0$ are fixed hyperparameters. See Appendix~\ref{sec:binary_concrete} for details about the Hard Concrete distribution including its parameterization. $\operatorname{NN}$ are feed-forward neural networks with architecture  $[H / 4, \tanh, 1]$ where $H$ is the BERT hidden size, $b$s are learnable biases. We use the same functional form to compute $z^{(\ell)}$ (masking hidden states) but $x_i$ omitted from the input of the feed-forward NN. For the input probe the output of the last projection (but not the bias) is constrained to be $\in (-\xi, \xi)$ for numerical stability. We initialized the bias of the last FFNN layer to $5$ to start with high probability of keeping states (fundamental for good convergence as the initialized \textsc{DiffMask} has not learned what to mask yet).

\section{The Hard Concrete distribution} \label{sec:binary_concrete}

The Hard Concrete distribution, assigns density to continuous outcomes in the open interval $(0,1)$ and non-zero mass to exactly $0$ and exactly $1$. A particularly appealing property of this distribution is that sampling can be done via a differentiable reparameterization~\citep{rezende2014stochastic, kingma2013auto}. 
In this way, the $\gL_0$ loss in Equation~\ref{eq:l0} becomes an expectation
\begin{equation} \label{eq:l0_relaxation}
	\gL_0(\phi,b|x) = \sum_{i=1}^N \E_{p_\phi(z_i|x)} \left[ z_i \neq 0 \right] \;.
\end{equation}
whose gradient can be estimated via Monte Carlo sampling without the need for REINFORCE and without introducing biases. We did modify the original Hard Concrete, though only so slightly, in a way that it gives support to samples in the half-open interval $[0, 1)$, that is,  with non-zero mass only at $0$. That is because we need only distinguish $0$ from non-zero, and the value $1$ is not particularly important.\footnote{Only a true $0$ is guaranteed to completely mask an input out, while any non-zero value, however small, may leak some amount of information.} 

\begin{figure}[t]
	\centering
	\begin{subfigure}[b]{0.23\textwidth}
		\centering
		\includegraphics[scale=0.27]{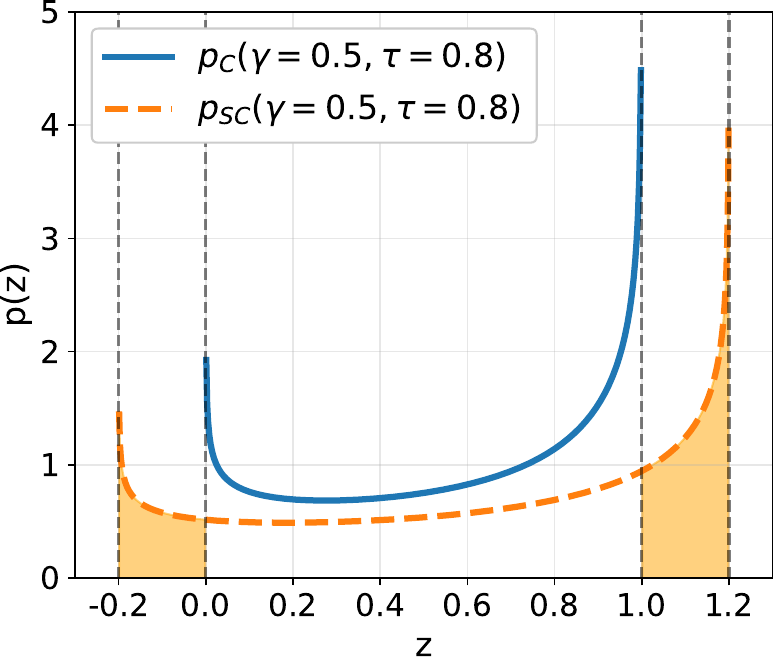}
		\caption{}
		\label{fig:concrete1}
	\end{subfigure}
	~
	\begin{subfigure}[b]{0.23\textwidth}
		\centering
		\includegraphics[scale=0.27]{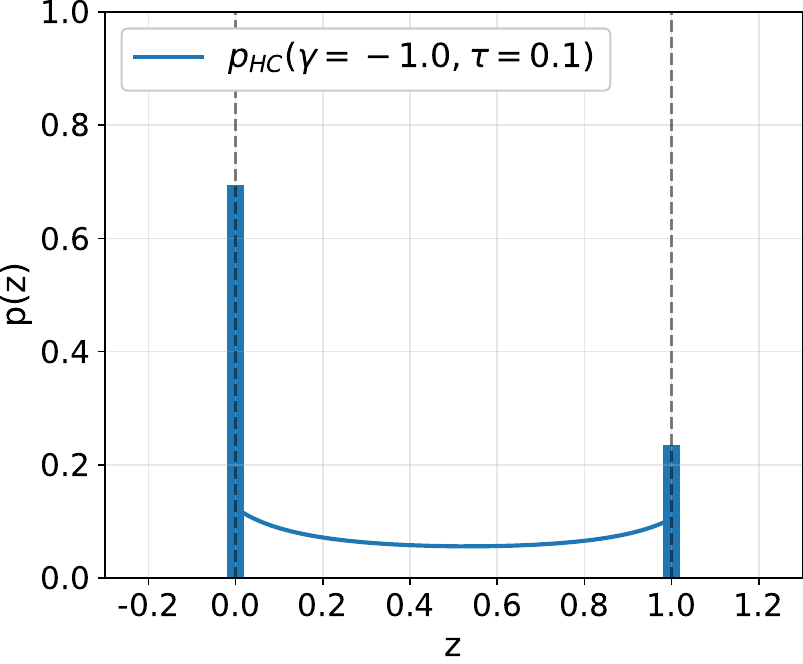}
		\caption{}
		\label{fig:concrete2}
	\end{subfigure}
	\caption{Binary Concrete distributions: (a) a Concrete $p_{C}$ and its stretched version $p_{SC}$; (b) a rectified and stretched (Hard) Concrete $p_{HC}$.}
	\label{fig:concrete}
\end{figure}

\paragraph{The distribution}
A stretched and rectified Binary Concrete (also known as Hard Concrete) distribution is obtained applying an affine transformation to the Binary Concrete distribution~\citep{maddison2016concrete, jang2016categorical} and rectifying its samples in the interval $[0,1]$ (see Figure~\ref{fig:concrete}). A Binary Concrete is defined over the open interval $(0, 1)$ ($p_{C}$ in Figure~\ref{fig:concrete1}) and it is parameterised by a location parameter $\gamma \in \R$ and temperature parameter $\tau \in \R_{>0}$. The location acts as a logit and it controls the probability mass skewing the distribution towards $0$ in case of negative location and towards $1$ in case of positive location. The temperature parameter controls the concentration of the distribution. The Binary Concrete is then stretched with an affine transformation extending its support to $(l, r)$ with $l \leq 0$ and $r \geq 1$ ($p_{SC}$ in Figure~\ref{fig:concrete1}). Finally, we obtain a Hard Concrete distribution rectifying samples in the interval $[0,1]$. This corresponds to collapsing the probability mass over the interval $(l, 0]$ to $0$, and the mass over the interval $[1, r)$ to $1$ ($p_{HC}$ in Figure~\ref{fig:concrete2}). This induces a distribution over the close interval $[0, 1]$ with non-zero mass at $0$ and $1$. Samples are obtained using
\begin{equation}
	\begin{aligned}
		s & = \sigma \left( \left(\log u - \log(1 - u) + \gamma  \right) / \tau  \right)   \;, \\
		z & = \min \left( 1, \max \left( 0, s \cdot \left( l - r \right) + r \right) \right) \;, 
	\end{aligned}
\end{equation}
where $\sigma$ is the Sigmoid function $\sigma(x) = (1 + e^{-x})^{-1}$ and $u\sim\gU(0,1)$. We point to the Appendix B of~\citet{louizos2017learning} for more information about the density of the resulting distribution and its cumulative density function.

\paragraph{Latent rationales} 
There is a stream of work on learning interpretable models by means of extracting latent rationales~\citep{lei-etal-2016-rationalizing,bastings2019interpretable}. Some of the techniques underlying \textsc{DiffMask} are related to that line of work, but overall we approach very different problems. \citet{lei-etal-2016-rationalizing} use REINFORCE to minimize a downstream loss computed on masked inputs, where the masks are binary and latent. They employ $L_0$ regularization to solve the task while conditioning only on small subsets of the input regarded as a \emph{rationale} for the prediction. To the same end,~\citet{bastings2019interpretable} minimize downstream loss subject to constraints on expected $L_0$ using a variant of the sparse relaxation of~\citet{louizos2017learning}. In sum, they employ stochastic masks to learn an interpretable model which they learn by minimizing a downstream loss subject to constraints on $L_0$, we employ stochastic masks to interpret an existing model and for that we minimize $L_0$ subject to constraints on that model's downstream performance.

\section{Hyperparameters} \label{sec:hyper}

\subsection{Toy task} \label{sec:hyper-toy}

\paragraph{Data}
We generate sequences of varying length (up to $10$ digits long) sampling each element independently: with $50\%$ probability, we draw uniformly $n$ or $m$ and, with $50\%$ probability, we draw uniformly from the remaining digits. We generate $10$k data-points, keeping $10\%$ of them for validation. The space of input sequences is $>\!10^{10}$. Thus, a model that solves the task cannot simply memorize the training set.

\paragraph{Model}
The precise model formulation is the following: given a query $q = \langle n, m \rangle$ and an input $x = \langle x_1, \dots x_t \rangle$, they are embedded as
\begin{equation}
	\begin{aligned}
        n' &= \operatorname{Emb}_q(n) \;,\\
        m' &= \operatorname{Emb}_q(m) \;,\\
        x_i' &= \operatorname{Emb}_x(x_i) & \forall i \in 1\dots t\;,
	\end{aligned}
\end{equation}
where $\operatorname{Emb}_q$ and $\operatorname{Emb}_x$ are embedding layers of dimensionality $64$. The prediction is computed as
\begin{equation}
	\begin{aligned}
        h_i^{(1)} &= \operatorname{FFNN}([n';m';x_i'])  & \forall i \in 1\dots t \;,\\
        h_0^{(2)} &= [0\dots 0]^\top  \;,\\
        h_i^{(2)} &= \operatorname{GRU}(h_{i}^{(1)}, h_{i-1}^{(2)})  & \forall i \in 1\dots t \;,\\
        y &= w^\top h_t^{(2)} + b \;,
	\end{aligned}
\end{equation}
where $[\cdot;\cdot]$ denotes concatenation, $\operatorname{FFNN}$ is a feed-forward neural network with architecture $[64\times 3, \tanh, 2]$, $\operatorname{GRU}$ is a Gated Recurrent Network~\citep{cho2014learning} with hidden size of $64$, and $w \in \R^{64}$, $b \in \R$ are the weight and bias parameter of the final classifier respectively.

\paragraph{Attribution methods}
Integrated gradient attribution~\citep{sundararajan2017axiomatic} is computed with $500$ steps. Attribution of~\citet{schulz2020restricting} is computed at token level with $\beta=10 / k$ where $k$ is the token embedding size. We optimized using the RMSprop~\citep{tieleman2012lecture} with learning rate $10^{-1}$ for $500$ steps. Attribution of~\citet{guan2019towards} is computed at token level with $\lambda=10^{-4}$ using RMSprop with learning rate $10^{-1}$ for $500$ steps. Our \textsc{DiffMask} is optimized for $100$ epochs using Lookahead RMSprop~\citep{tieleman2012lecture,zhang2019lookahead} with learning rate $10^{-2}$ for $\phi,b$ and $10^{-1}$ for $\alpha$. For these attribution methods we used our own re-implementation.

\begin{table}[t]
    \centering
    \begin{tabular}{l|l}
    \toprule
    \textbf{Model} & \textbf{Value} \\
    \midrule
    Type                & BERT\textsubscript{BASE} (uncased) \\
    Layers              & $12$ \\
    Hidden units        & $768$ \\
    Pre-trained masking & standard \\
    Optimizer           & Adam* \\
    Learning rate       & $3 \cdot 10^{-5}$ \\
    Train epochs        & $50$ \\
    Batch size          & $64$ \\
    \midrule
    \textbf{\textsc{DiffMask}} & \textbf{Value} \\
    \midrule
    Optimizer           & Lookahead RMSprop** \\
    Learning rate $\phi,b$      & $3 \cdot 10^{-4}$ \\
    Learning rate $\lambda$     & $1 \cdot 10^{-1}$ \\
    Train epochs        & $100$ \\
    Batch size          & $64$ \\
    Constrain           & $\mathrm{D_{KL}}[y \| \hat y] < 0.5$ \\
    \bottomrule
    \end{tabular}
    \caption{Hyperparameters for the sentiment classification experiment. Optimizers: *~\citet{kingma2014adam}, **~\citet{tieleman2012lecture,zhang2019lookahead}.}
    \label{tab:hyper-sst}
\end{table}

\subsection{Sentiment Classification} \label{sec:hyper-sst}

\paragraph{Data}
We used the Stanford Sentiment Treebank ~\citep[SST;][]{socher2013recursive} available here\footnote{\url{https://nlp.stanford.edu/sentiment/trainDevTestTrees_PTB.zip}}. We pre-processed the data as in~\citet{bastings2019interpretable}. Training and validation sets contain $8544$ and $1101$ sentences respectively.

\paragraph{Model}
For the sentiment classification experiment we downloaded\footnote{\url{https://huggingface.co/transformers/pretrained_models.html}\label{note:huggingface-model}} a pre-trained model from the Huggingface implementation\footnote{\url{https://github.com/huggingface/transformers}\label{note:huggingface-repo}} of~\citet{wolf2019huggingface}, and we fined-tuned on the SST dataset. We report hyperparameters used for training the model and our \textsc{DiffMask} in Table~\ref{tab:hyper-sst}.

\subsection{Question Answering} \label{sec:hyper-qa}

\paragraph{Data}
We used the Stanford Question Answering Dataset~\citep[\textsc{SQuAD v1.1};][]{rajpurkar2016squad} available here\footnote{\url{https://rajpurkar.github.io/SQuAD-explorer}}.
Pre-processing excluded QA pairs with more than $384$ BPE tokens to avoid memory issues. After this we end up having $86706$ training instances and $10387$ validation instances.

\paragraph{Model}
For the question answering experiment we downloaded~\footref{note:huggingface-model} an already fine-tuned model from the Huggingface implementation\footref{note:huggingface-repo} of~\citet{wolf2019huggingface} We report hyperparameters used by them for training the original model and the ones used for our \textsc{DiffMask} in Table~\ref{tab:hyper-qa}.

\begin{table}[t]
    \centering
    \begin{tabular}{l|l}
    \toprule
    \textbf{Model} & \textbf{Value} \\
    \midrule
    Type                & BERT\textsubscript{LARGE} (uncased) \\
    Layers              & $24$ \\
    Hidden units        & $1024$ \\
    Pre-trained masking & whole-word \\
    Optimizer           & Adam* \\
    Learning rate       & $3 \cdot 10^{-5}$ \\
    Train epochs        & $2$ \\
    Batch size          & $24$ \\
    \midrule
    \textbf{\textsc{DiffMask}} & \textbf{Value} \\
    \midrule
    Optimizer           & Lookahead RMSprop** \\
    Learning rate $\phi,b$       & $3 \cdot 10^{-4}$ \\
    Learning rate $\lambda$      & $1 \cdot 10^{-1}$ \\
    Epochs (inputs)       & $1$ (per layer) \\
    Epochs (hidden)       & $4$ \\
    Batch size          & $8$ \\
    Constrain           & $\mathrm{D_{KL}}[y \| \hat y] < 1$ \\
    \bottomrule
    \end{tabular}
    \caption{Hyperparameters for the question answering experiment. Optimizers: *~\citet{kingma2014adam}, **~\citet{tieleman2012lecture,zhang2019lookahead}.}
    \label{tab:hyper-qa}
\end{table}

\section{Additional plots and results} \label{sec:extra}
\begin{figure}[t]
	\centering
	\includegraphics[width=0.45\textwidth]{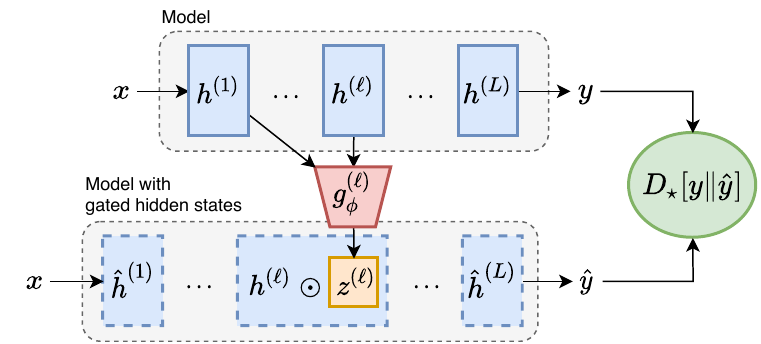}
	\caption{\textsc{DiffMask} for hidden states: states up to layer $\ell$ from a model (top) are fed to a classifier $g$ that predicts a mask $z^{(\ell)}$. We use this to mask the $\ell$ih hidden state and re-compute the forward pass from that point on (bottom). The classifier $g$ is trained to mask the hidden state as much as possible without changing the output (minimizing a divergence $\mathrm{D_\star}$).}
	\label{fig:model-hidden}
\end{figure}

In Figure~\ref{fig:model-hidden} we show an overview of the variant of \textsc{DiffMask} to analyze the hidden states of a model~(see Figure~\ref{fig:model-input} to compare the two versions).

\subsection{Toy task} \label{sec:toy-extra}
\begin{figure}[t]
    \centering
    \includegraphics[width=.3\textwidth]{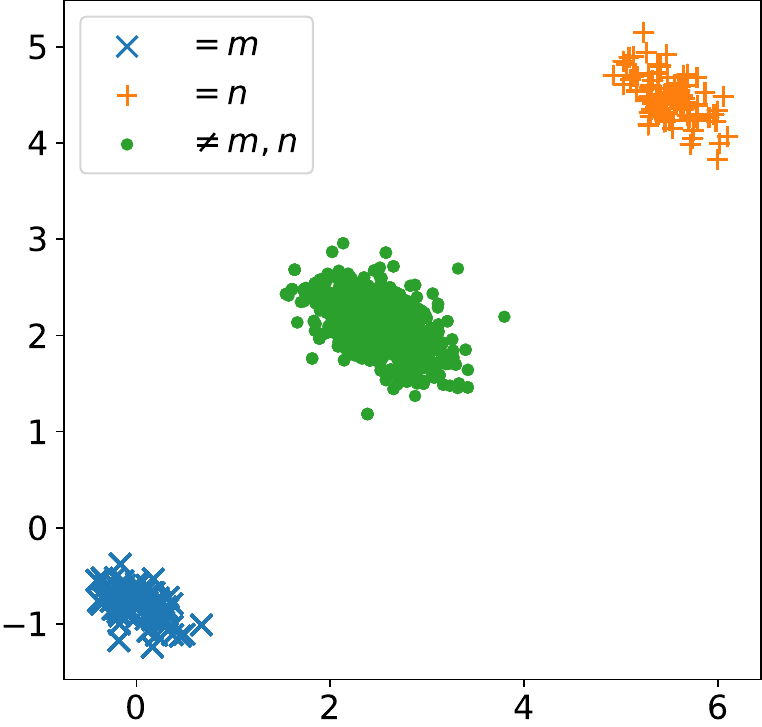}
    \caption{Hidden state values for the two-neuron toy task. Clusters of whether the input digit is equal to the first or second position in the query ($=n$ or $=m$ respectively) or not at all ($\neq n,m$) are completely linear separable.}
    \label{fig:toy-hidden-projection}
\end{figure}

In Figure~\ref{fig:toy-hidden-projection} we show the distribution of hidden states in the toy task where we highlight whether they belong to a state corresponding to $n,m$ or neither of them.

\subsection{Sentiment Classification} \label{sec:sst-extra}
\begin{figure}[t]
	\centering
	\begin{subfigure}[b]{0.23\textwidth}
		\centering
		\includegraphics[scale=.3]{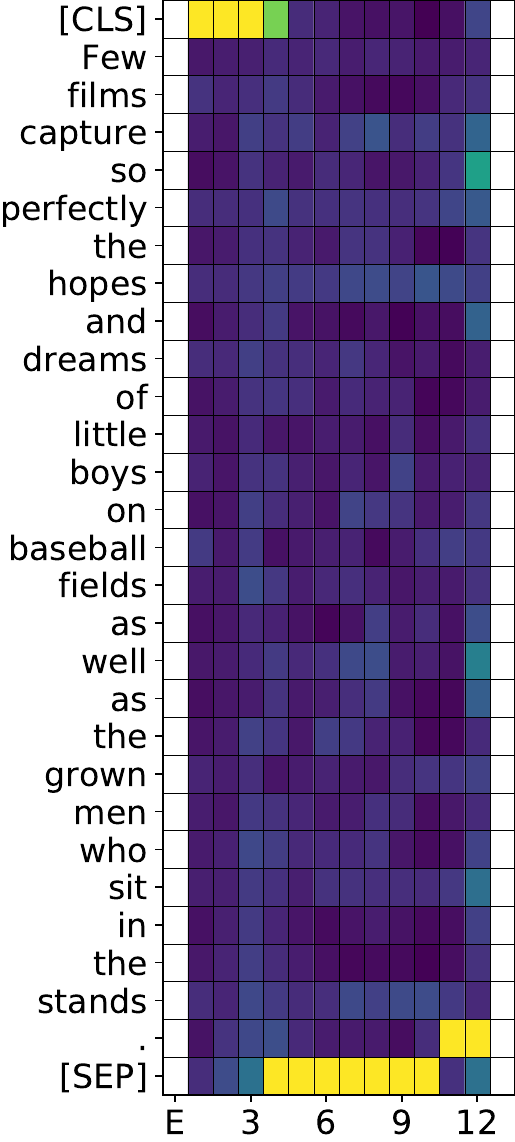}
		\caption{Attention.}
		\label{fig:sst-att-2}
	\end{subfigure}
	~
	\begin{subfigure}[b]{0.23\textwidth}
		\centering
		\includegraphics[scale=.3]{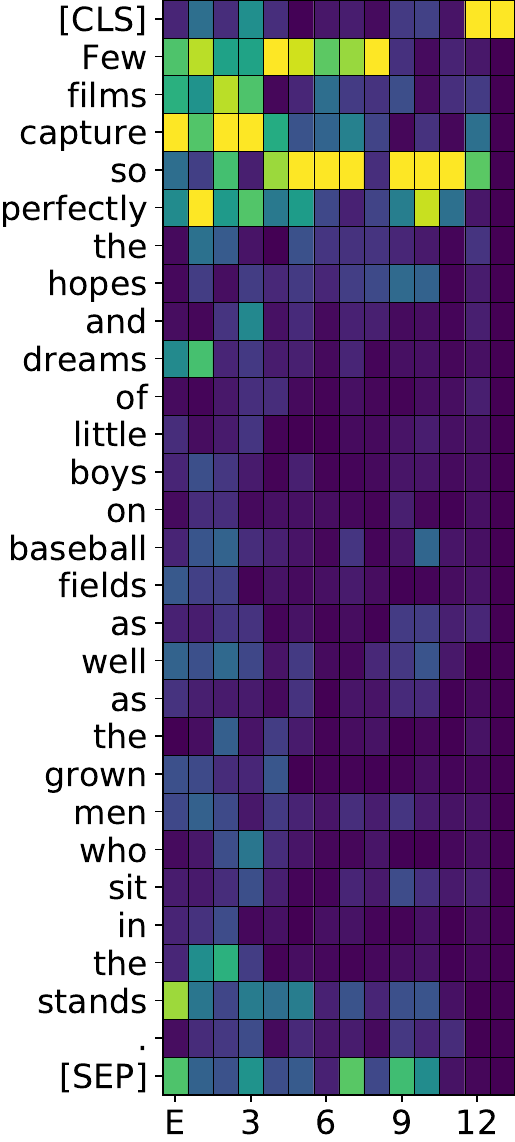}
		\caption{\citet{sundararajan2017axiomatic}.}
		\label{fig:sst-ig2}
	\end{subfigure}
    \par\vspace{6pt}
	\begin{subfigure}[b]{0.23\textwidth}
		\centering
		\includegraphics[scale=.3]{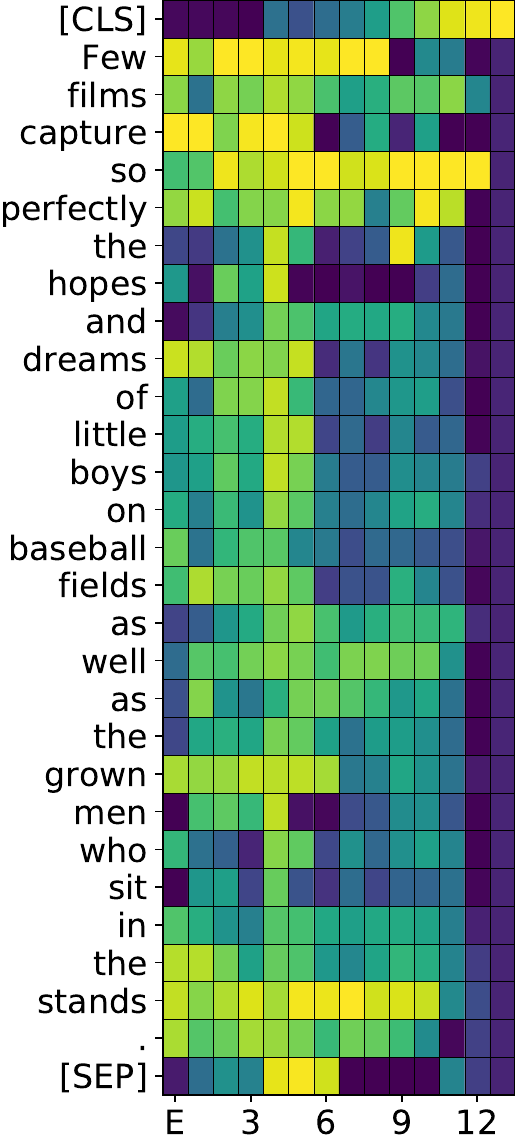}
		\caption{\citet{schulz2020restricting}.}
		\label{fig:sst-schulz2}
	\end{subfigure}
	~
	\begin{subfigure}[b]{0.23\textwidth}
		\centering
		\includegraphics[scale=.3]{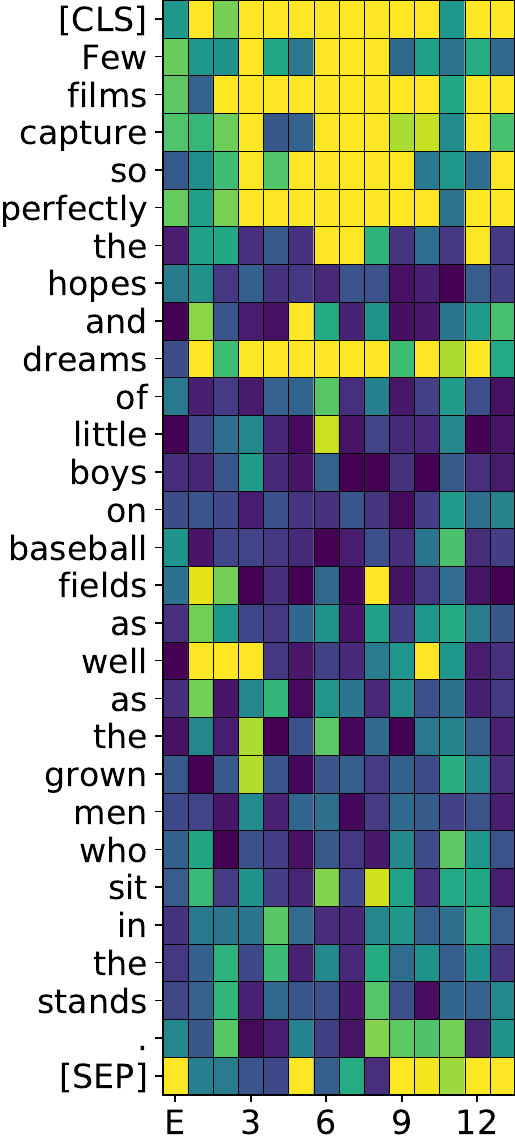}
		\caption{\citet{guan2019towards}.}
		\label{fig:sst-guan2}
	\end{subfigure}
    \par\vspace{6pt}
	\begin{subfigure}[b]{0.23\textwidth}
		\centering
		\includegraphics[scale=.3]{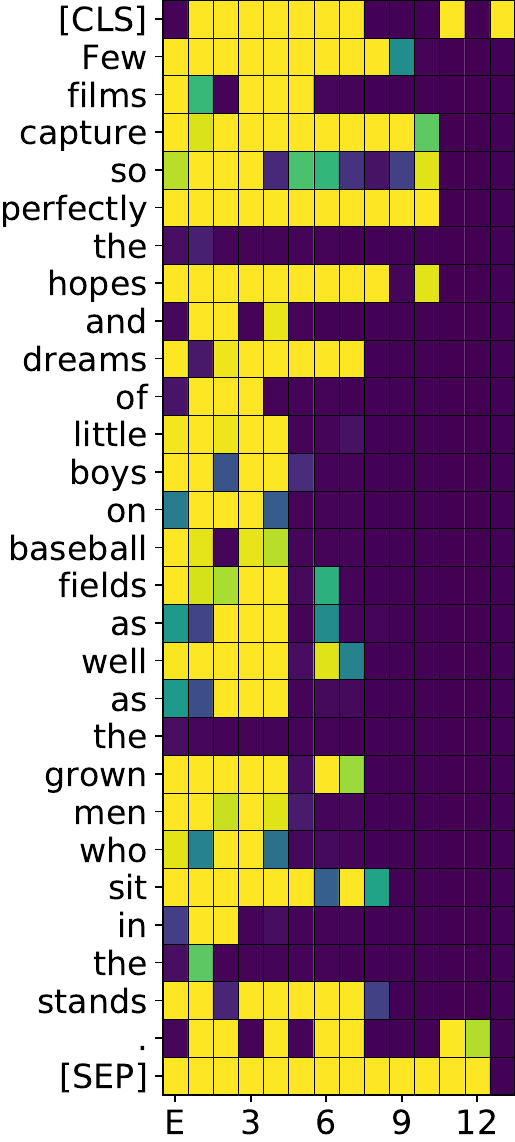}
		\caption{\textsc{DiffMask}.}
		\label{fig:sst-ours-hidden2}
	\end{subfigure}
	~
	\begin{subfigure}[b]{0.23\textwidth}
		\centering
		\includegraphics[scale=.3]{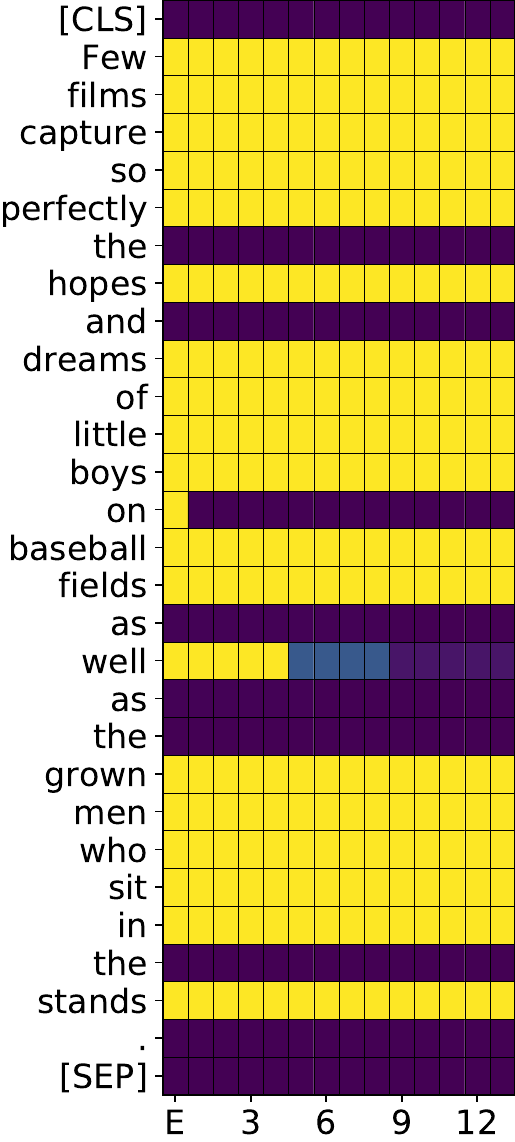}
		\caption{\textsc{DiffMask} on input.}
		\label{fig:sst-ours-input2}
	\end{subfigure}
	\caption{Sentiment classification: comparison between attribution method for hidden layers w.r.t. the predicted label. All plots are normalized per-layer by the largest attribution. Attention heatmap is obtained max pooling over heads and averaging across positions.}
	\label{fig:sst-comparison2}
\end{figure}

In Figure~\ref{fig:sst-comparison2} we show an additional comparison example between attribution method for hidden layers w.r.t the predicted label.

\subsubsection{Ablation} \label{sec:sst-ablations}
\begin{figure}[t]
	\centering
	\begin{subfigure}[t]{0.45\textwidth}
		\centering
		\includegraphics[scale=.3]{plots/sst-hidden-46-diffmask.pdf}
		\caption{Masking hidden states with amortization.}
		\label{fig:sst-ablation-normal}
	\end{subfigure}
    \par\vspace{6pt}
	\begin{subfigure}[t]{0.45\textwidth}
		\centering
		\includegraphics[scale=.3]{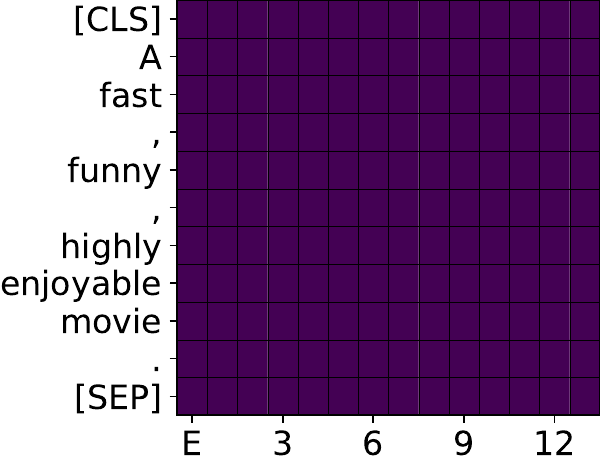}
		\caption{Masking hidden states without amortization.}
		\label{fig:sst-ablation-noamo}
	\end{subfigure}
	\par\vspace{6pt}
	\begin{subfigure}[t]{0.45\textwidth}
		\centering
		\includegraphics[scale=.3]{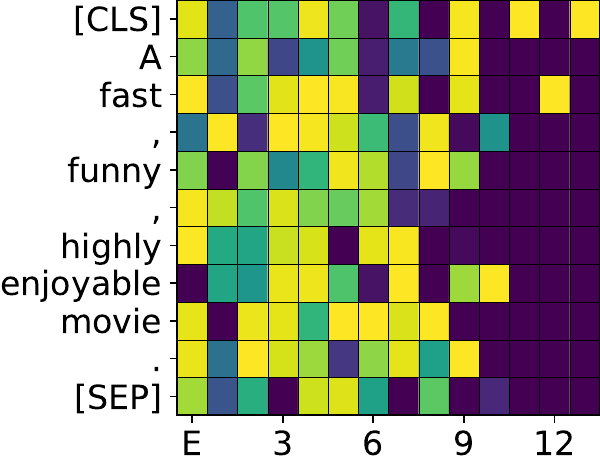}
		\caption{Masking hidden states without amortization and without baseline.}
		\label{fig:sst-ablation-noamo-nobase}
	\end{subfigure}
	\caption{Sentiment classification: ablation study on amortization and baseline.}
	\label{fig:sst-ablation}
\end{figure}

As argued in the introduction and shown on the toy task, many popular methods (e.g., erasure and its approximations) are over-aggressive in discarding inputs and hidden units. Amortization is a fundamental component of \textsc{DiffMask} and is aimed at addressing this issue. In Figure~\ref{fig:sst-ablation} we show how our method behaves when ablating amortization and thus optimizing on a single example instead. Noticeable, our method converges to masking out all hidden states at any layer (Figure~\ref{fig:sst-ablation-noamo}). This happens as it learns an \textit{ad hoc} baseline just for that example. When we ablate both amortization and baseline learning (Figure~\ref{fig:sst-ablation-noamo-nobase}), the method struggles to uncover any meaningful patterns. This highlights how both core components of our method are needed in combination with each other.

\subsection{Question Answering} \label{sec:squad-extra}
\begin{figure}[t]
	\centering
	\begin{subfigure}[b]{0.23\textwidth}
		\centering
		\includegraphics[width=\textwidth]{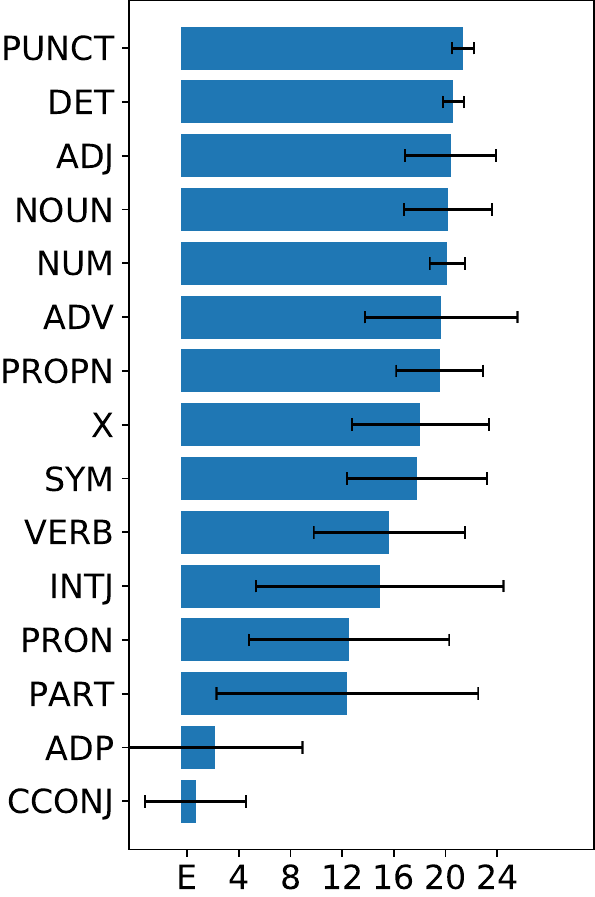}
		\caption{Question inputs.}
		\label{fig:squad-stats-pos-input-question}
	\end{subfigure}
	~
	\begin{subfigure}[b]{0.23\textwidth}
		\centering
		\includegraphics[width=\textwidth]{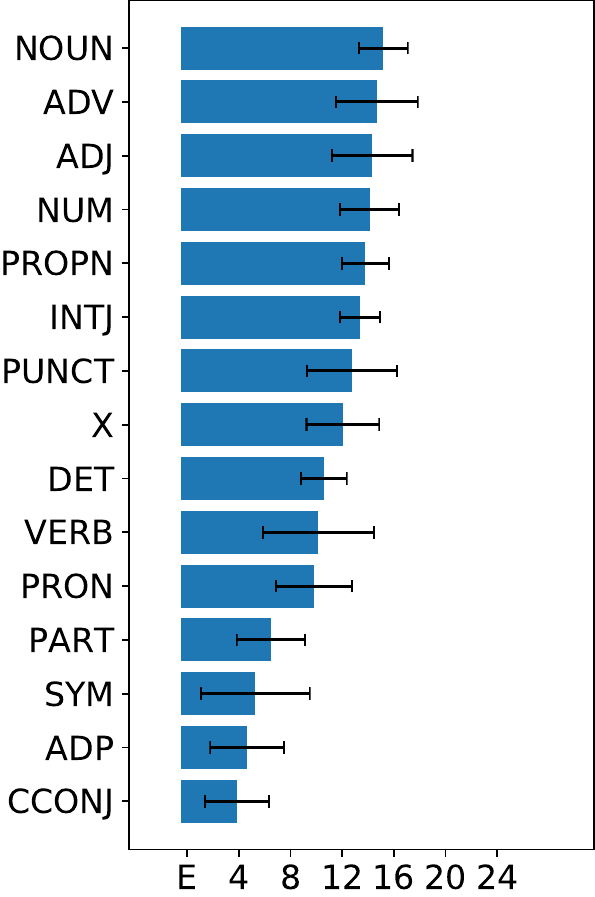}
		\caption{Question hidden states.}
		\label{fig:squad-stats-pos-hidden-question}
	\end{subfigure}
    \par\vspace{6pt}
	\begin{subfigure}[b]{0.23\textwidth}
		\centering
		\includegraphics[width=\textwidth]{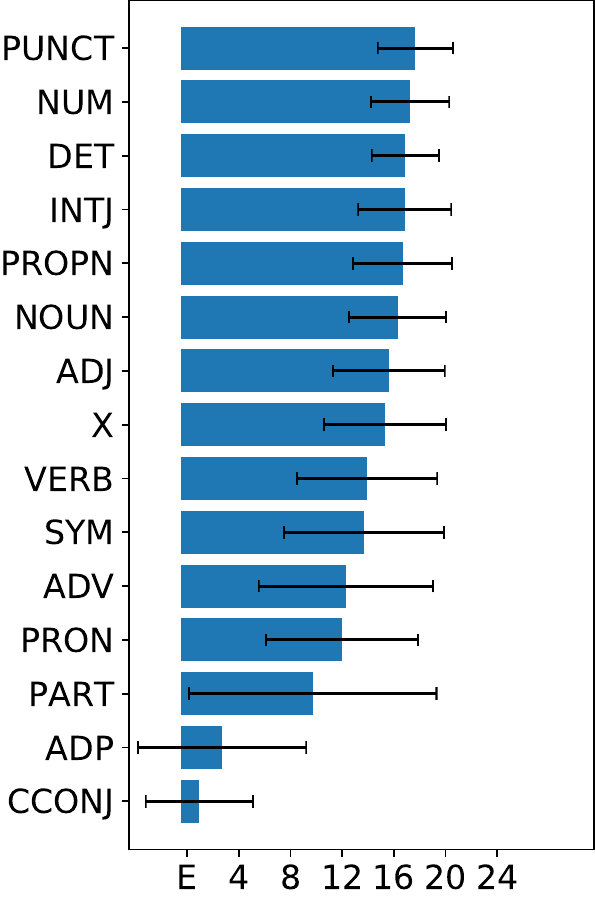}
		\caption{Context inputs.}
		\label{fig:squad-stats-pos-input-context}
	\end{subfigure}
	~
	\begin{subfigure}[b]{0.23\textwidth}
		\centering
		\includegraphics[width=\textwidth]{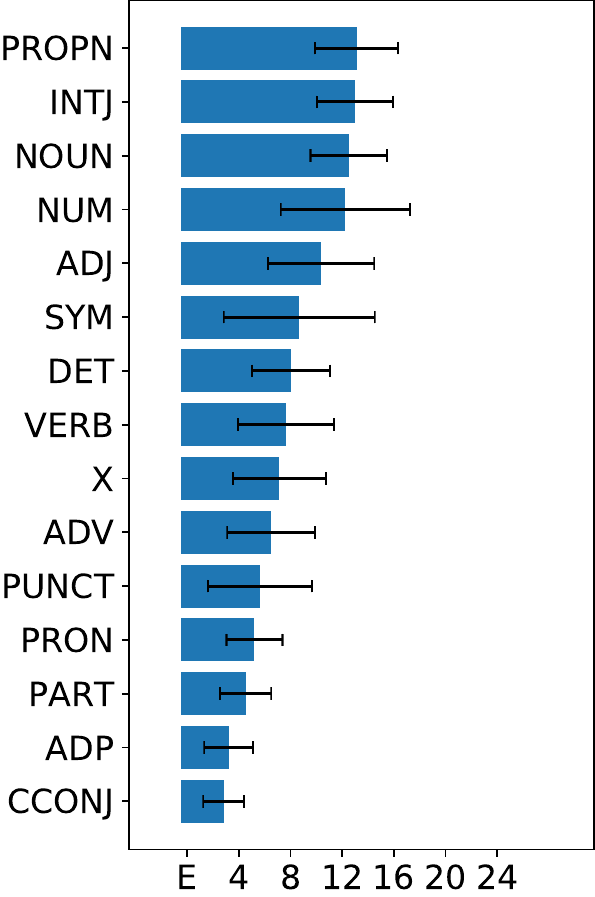}
		\caption{Context hidden states.}
		\label{fig:squad-stats-pos-hidden-context}
	\end{subfigure}
    \par\vspace{6pt}
	\begin{subfigure}[b]{0.23\textwidth}
		\centering
		\includegraphics[width=\textwidth]{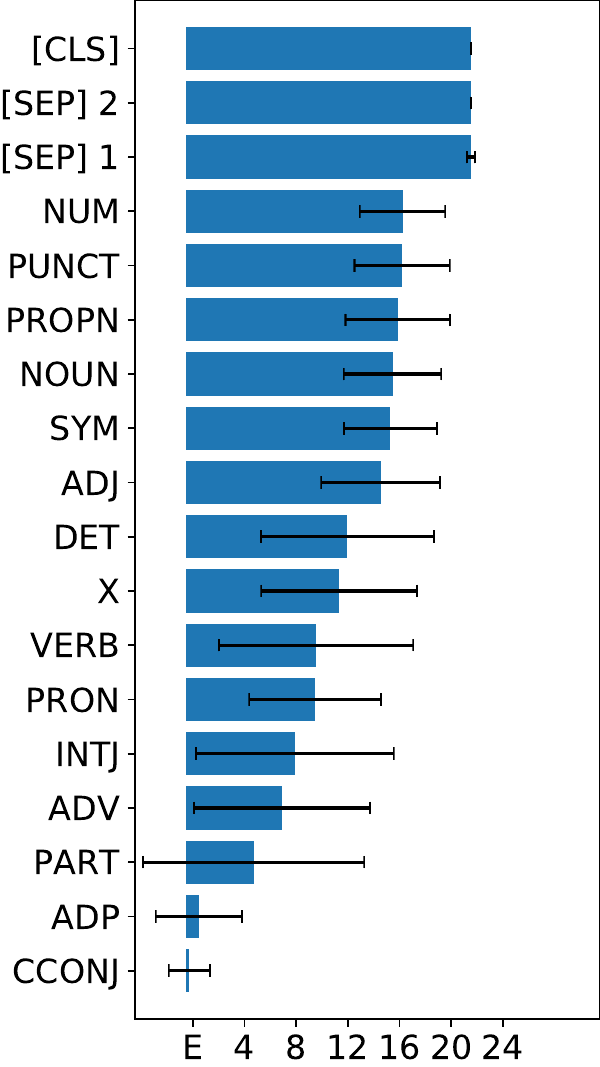}
		\caption{All inputs.}
		\label{fig:squad-stats-pos-input-all}
	\end{subfigure}
	~
	\begin{subfigure}[b]{0.23\textwidth}
		\centering
		\includegraphics[width=\textwidth]{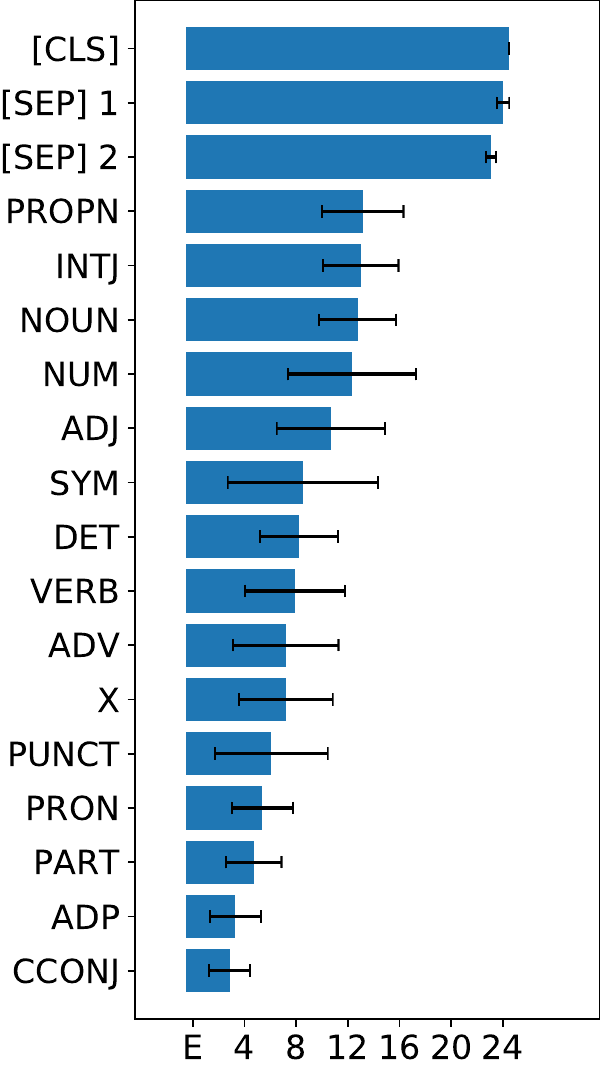}
		\caption{All hidden states.}
		\label{fig:squad-stats-pos-hidden-all}
	\end{subfigure}
	\caption{Question answering: average number of layers that predict to keep input tokens (a), (c) and (e) or hidden states (b), (d) and (f) aggregating by part-of-speech tag (POS) on validation set.}
	\label{fig:squad-stat-pos}
\end{figure}

\begin{figure}[t]
    \centering
	\begin{subfigure}[b]{0.23\textwidth}
		\centering
		\includegraphics[scale=0.28]{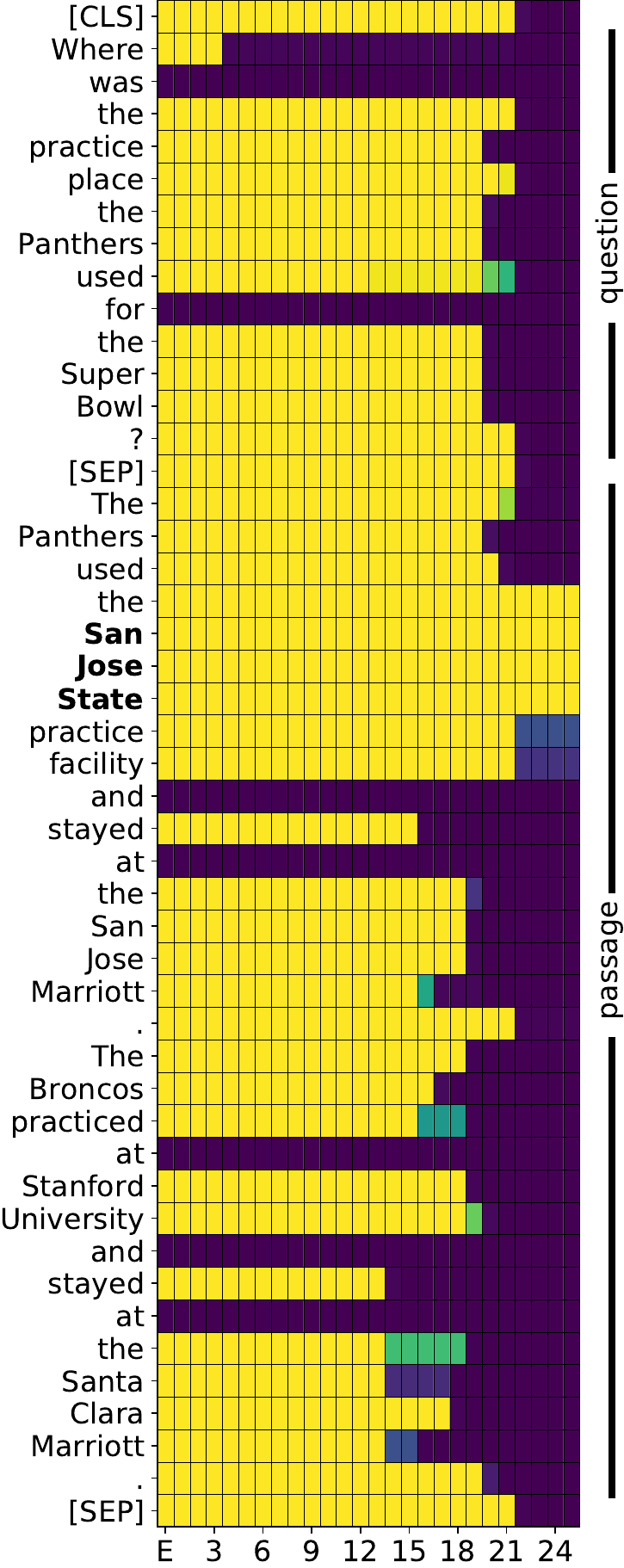}
		\caption{Gating the input.}
		\label{fig:squad-visual-input2}
	\end{subfigure}
	~
	\begin{subfigure}[b]{0.23\textwidth}
		\centering
		\includegraphics[scale=0.28]{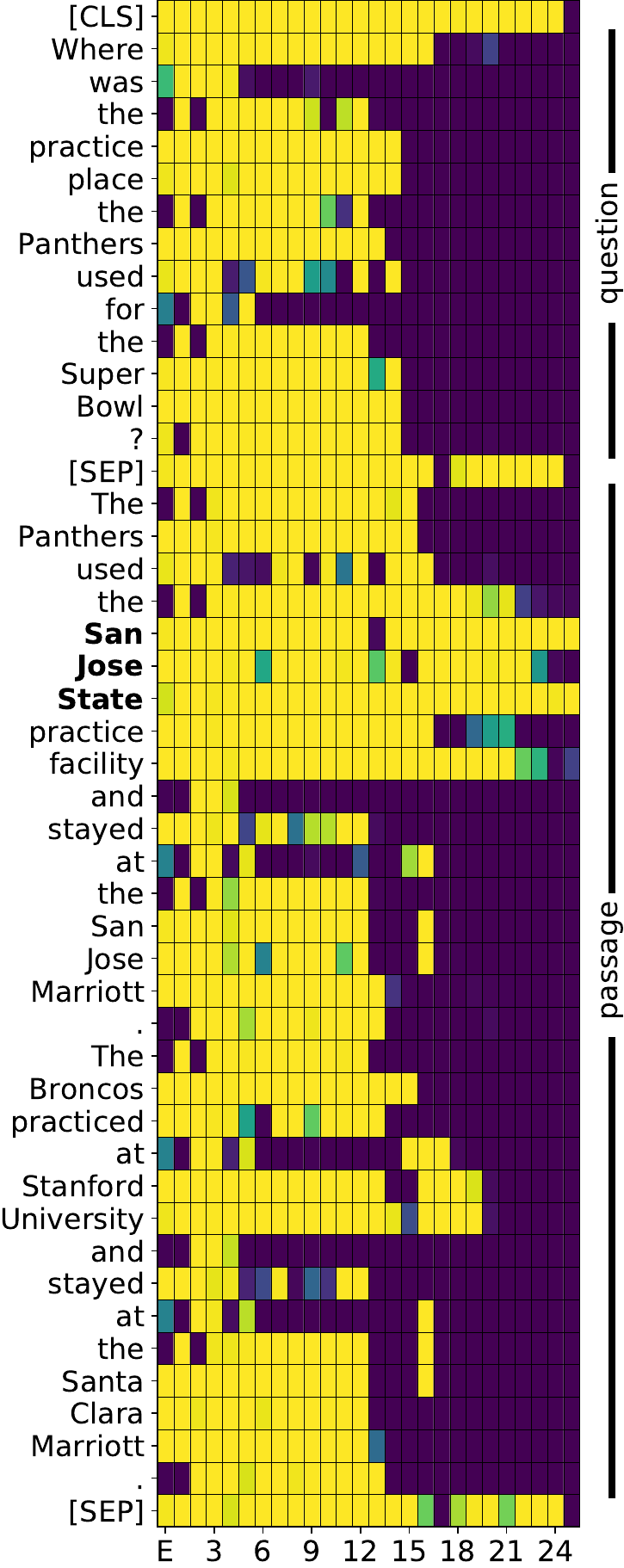}
		\caption{Gating hidden states.}
		\label{fig:squad-visual-hidden2}
	\end{subfigure}
	\par\vspace{6pt}
	\begin{subfigure}[b]{0.23\textwidth}
		\centering
		\includegraphics[scale=0.28]{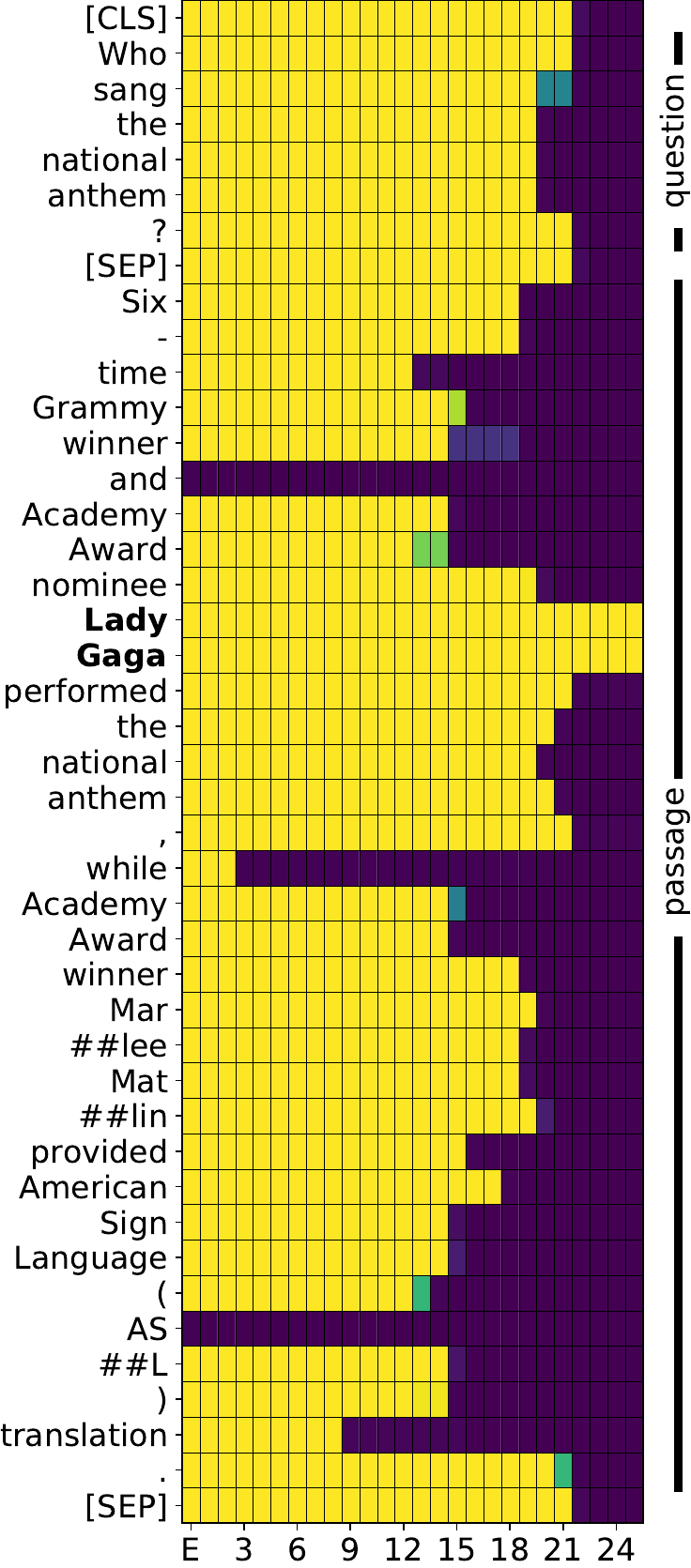}
		\caption{Gating the input.}
		\label{fig:squad-visual-input3}
	\end{subfigure}
	~
	\begin{subfigure}[b]{0.23\textwidth}
		\centering
		\includegraphics[scale=0.28]{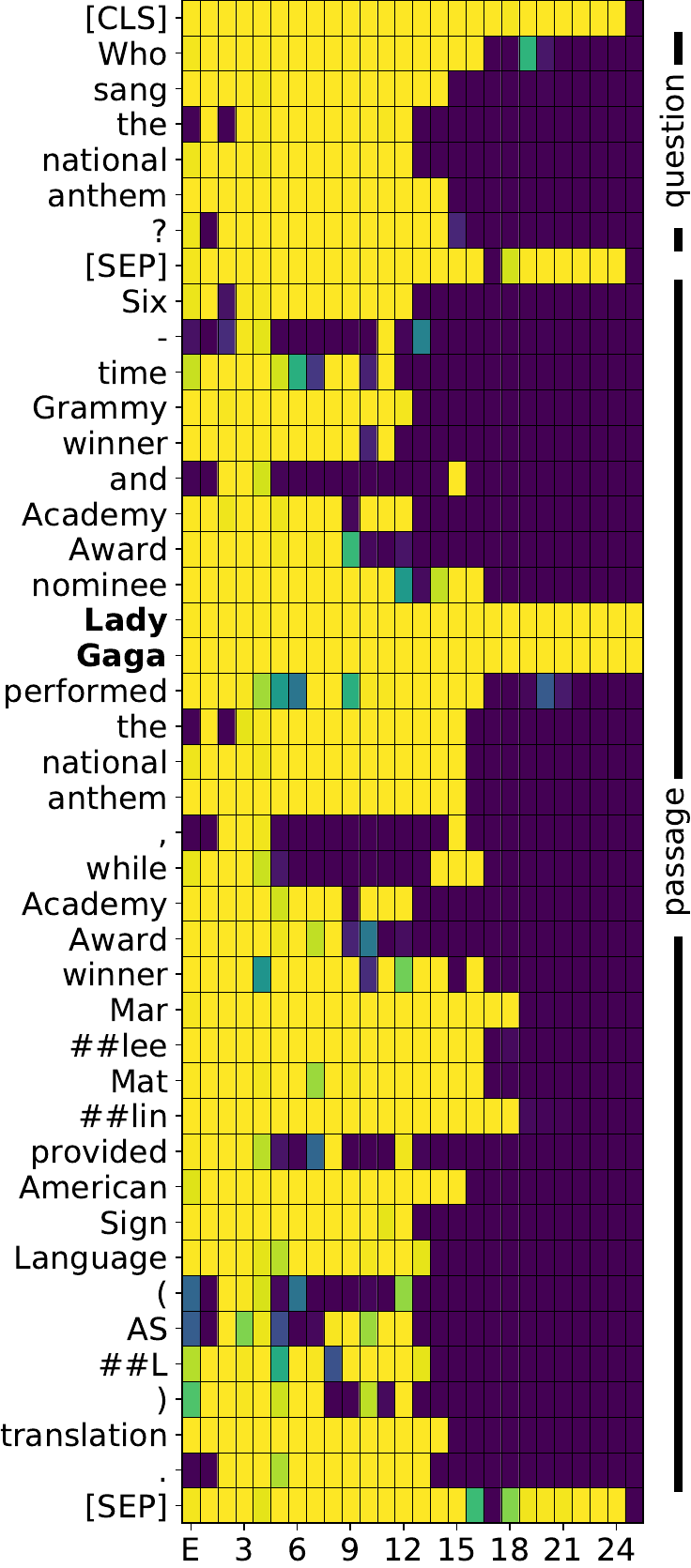}
		\caption{Gating hidden states.}
		\label{fig:squad-visual-hidden3}
	\end{subfigure}
	\caption{Expectation predicted by \textsc{DiffMask} to keep the inputs in (a) (c) and hidden states in (b) (d) on two different QA pairs. The correct answers is highlighted in bold.}
	\label{fig:squad-visual2}
\end{figure}

In Figure~\ref{fig:squad-stat-pos} we report statistics on the average number of layers that predict to keep input tokens aggregating by POS tag. We report additional two examples of expectation predicted by \textsc{DiffMask} in Figure~\ref{fig:squad-visual2}.

\end{document}